\documentclass[a4paper,fleqn]{cas-dc}

\usepackage[numbers]{natbib}

\usepackage{pifont}
\newcommand{\cmark}{\ding{51}}
\newcommand{\xmark}{\ding{55}}
\usepackage{xcolor}
\definecolor{myblue}{HTML}{3F99E2}
\definecolor{mygreen}{HTML}{009900}

\begin{document}
\let\WriteBookmarks\relax
\def\floatpagepagefraction{1}
\def\textpagefraction{.001}
% \shorttitle{GeoVista: Visually Grounded Active Perception for Vision-Language Understanding of Ultra-High-Resolution Remote Sensing Images}
\shortauthors{Ronghao Fu et~al.}

\title [mode = title]{GeoVista: Visually Grounded Active Perception for Vision-Language Understanding of Ultra-High-Resolution Remote Sensing Images}

\author[1,2]{Jiashun Zhu}[style=chinese]
% \cormark[1]
\ead{zhujs25@mails.jlu.edu.cn}
\credit{Conceptualization, Methodology, Software, Validation, Formal analysis, Investigation, Writing - Original Draft, Visualization}

\author[1,2]{Ronghao Fu}[style=chinese]
\cormark[2]
\ead{furh@jlu.edu.cn}
\credit{Conceptualization, Methodology, Software, Validation, Formal analysis, Investigation, Writing - Original Draft, Visualization}

\author[1,2]{Jiasen Hu}[style=chinese]
\ead{hujs5522@mails.jlu.edu.cn}
\credit{Conceptualization, Methodology, Software, Validation, Formal analysis, Investigation, Writing - Original Draft, Visualization}

\author[3,4]{Jing Huang}[style=chinese]
\ead{huangjing@iie.ac.cn}
\credit{Conceptualization, Methodology, Software, Validation, Formal analysis, Investigation, Writing - Original Draft, Visualization}

\author[1,2]{Nachuan Xing}[style=chinese]
\ead{xingnc25@mails.jlu.edu.cn}
\credit{Conceptualization, Methodology, Software, Validation, Formal analysis, Investigation, Writing - Original Draft, Visualization}

\author[1,2]{Bo Yang}[style=chinese]
\cormark[2]
\ead{ybo@jlu.edu.cn}
\credit{Conceptualization, Methodology, Software, Validation, Formal analysis, Investigation, Writing - Original Draft, Visualization}

\address[1]{College of Computer Science and Technology, Jilin University, Changchun 130021, China}
\address[2]{Key Laboratory of Symbolic Computation and Knowledge Engineering of Ministry of Education, Changchun 130021, China}
\address[3]{Institute of Information Engineering, Chinese Academy of Sciences, Beijing 100000, China}
\address[4]{School of Cyber Security, University of Chinese Academy of Sciences, Beijing 100000, China}

% \cortext[cor1]{Equal contribution}
\cortext[cor2]{Corresponding author}

\begin{abstract}
Interpreting ultra-high-resolution (UHR) remote sensing images requires models to search for sparse and tiny visual evidence across large-scale scenes. Existing remote sensing vision-language models can inspect local regions with zooming and cropping tools, but most exploration strategies follow either a one-shot focus or a single sequential trajectory. Such single-path exploration can lose global context, leave scattered regions unvisited, and revisit or count the same evidence multiple times. To this end, we propose GeoVista, a planning-driven active perception framework for UHR remote sensing interpretation. Instead of committing to one zooming path, GeoVista first builds a global exploration plan, then verifies multiple candidate regions through branch-wise local inspection, while maintaining an explicit evidence state for cross-region aggregation and de-duplication. To enable this behavior, we introduce APE-GRO, a cold-start supervised trajectory corpus that reformulates diverse UHR tasks as Global-Region-Object interactive reasoning processes with a unified, scale-invariant spatial representation. We further design an Observe-Plan-Track mechanism for global observation, adaptive region inspection, and evidence tracking, and align the model with a GRPO-based strategy using step-wise rewards for planning, localization, and final answer correctness. Experiments on RSHR-Bench, XLRS-Bench, and LRS-VQA show that GeoVista achieves state-of-the-art performance. Code and dataset are available at \url{https://github.com/ryan6073/GeoVista}.
\end{abstract}


\maketitle

\section{Introduction}
Vision-language models (VLMs) are becoming an important interface between remote sensing (RS) imagery and human-centered Earth observation. Visual sensing captures spatial structures and physical patterns on the Earth's surface, while language enables users to query, describe, and reason about RS scenes in an interpretable form. Recent remote sensing vision-language models (RS-VLMs) extend this interface beyond fixed-category recognition by supporting tasks such as visual question answering, visual grounding, image captioning, and scene interpretation. These capabilities make RS-VLMs a promising foundation for flexible and task-adaptive geographic understanding.

A central difficulty for RS-VLMs lies in ultra-high-resolution (UHR) RS imagery. In practical satellite products, a single image may cover a very large geographic area while preserving fine spatial details. For example, Jilin-1 wide-swath satellites can provide panchromatic imagery with 0.5 m spatial resolution and a swath width of more than 150 km, corresponding to a cross-track image dimension of approximately 300,000 pixels. Compared with natural images, such data contain larger backgrounds, more sparsely distributed targets, and stronger scale variation. Many UHR tasks therefore cannot be solved by recognizing a single centered object. Such tasks often require the model to determine where task-relevant evidence may appear, inspect local details, and aggregate evidence across distant regions.

Existing methods approach this problem from three directions. High-resolution or dynamic-resolution encoding increases the amount of visual information available to VLMs, but larger inputs also introduce substantial background redundancy and do not indicate which regions should be inspected more carefully. Token pruning, visual compression, and coarse-to-fine selection methods, including text-guided pruning, GeoLLaVA-8K~\cite{Wang2025GeoLLaVA8KSR}, and C2F~\cite{Luo2025WhenLV}, improve efficiency by retaining compact visual evidence. However, because selection is usually performed before reasoning has fully unfolded, missed small objects or alternative candidate regions are difficult to recover later. Crop- and zoom-based methods, such as ZoomEarth~\cite{liu2025zoomearth}, ZoomSearch~\cite{ZoomSearch}, and GeoEyes~\cite{wang2026geoeyes}, introduce local inspection at inference time, but their search processes are commonly organized as either one-shot focusing or a single sequential zooming trajectory.

\begin{figure}
    \centering
    \includegraphics[width=\linewidth]{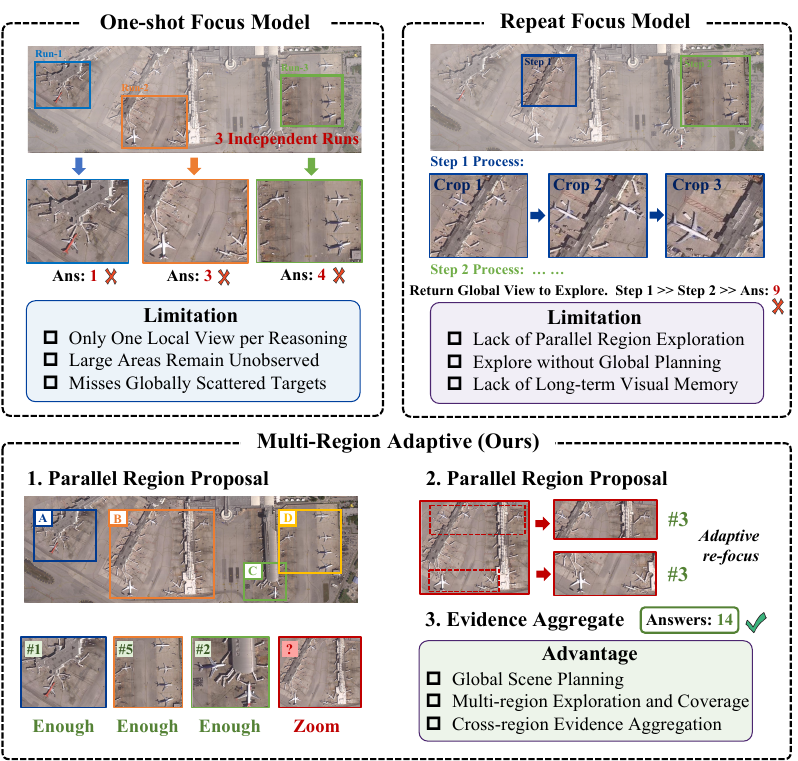}
    \caption{Comparison of three active perception paradigms for UHR remote sensing. \textbf{Left:} single-focus one-shot cropping misses scattered targets. \textbf{Middle:} repeated-focus sequential zooming lacks global planning and evidence memory. \textbf{Right:} the proposed multi-region active perception paradigm plans multiple ROIs, verifies them branch-wise, and aggregates cross-region evidence.}
    \label{fig:motivation}
\end{figure}

These designs leave a gap between visual access and evidence acquisition. UHR reasoning often requires the model to compare multiple spatially separated regions, verify uncertain candidates, count distributed targets, and avoid using the same evidence repeatedly. As illustrated in Fig.~\ref{fig:motivation}, single-focus methods may enlarge one local region while leaving other relevant areas unexamined. Sequential zooming methods allow multi-step observation, but the process tends to follow a depth-first path toward one candidate. Once the model commits to an incomplete or incorrect path, alternative regions may remain unexplored or insufficiently verified. Without an explicit evidence state, the model also has limited ability to remember which regions have been inspected, which findings have been confirmed, and which sub-goals remain unresolved.

\begin{figure}
    \centering
    \includegraphics[width=\columnwidth]{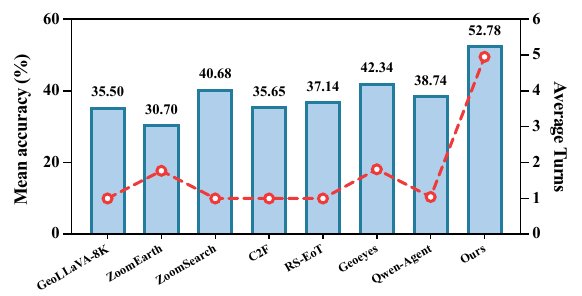}
    \caption{Performance comparison of different models on XLRS-Bench. The blue bars represent the mean accuracy, while the red dashed line indicates the average turns required for each method.}
    \label{fig:statistics}
\end{figure}

We refer to this failure mode as \emph{single-trajectory bias}: the tendency of an active perception model to acquire visual evidence along a narrow observation path without maintaining multiple candidate regions and their inspection states. Fig.~\ref{fig:statistics} further suggests that current active focusing strategies do not yet fully exploit multi-step visual evidence acquisition. Although observation tools are available, existing methods invoke them only a limited number of times on average and still leave a clear performance gap on XLRS-Bench. This motivates a more structured form of active perception for UHR remote sensing: an RS-VLM should preserve global context, plan multiple local inspections, track the evidence collected from each region, and decide when sufficient evidence has been gathered.

In this paper, we introduce \textbf{GeoVista}, a planning-driven active perception framework for multi-region exploration over UHR RS imagery. GeoVista is built around three design requirements derived from the above analysis. First, the model needs cold-start supervision that teaches it how to search rather than only how to answer. To this end, we construct \textbf{APE-GRO}, a tool-interleaved trajectory dataset that reformulates diverse UHR tasks as Global-Region-Object reasoning processes. With a unified scale-invariant spatial representation, APE-GRO supervises how to start from a global scene, identify regional candidates, and verify object-level evidence across observation scales. Second, the model needs an inference mechanism that can avoid single-path commitment. We therefore design an \textbf{Observe-Plan-Track} mechanism in which GeoVista observes the global view, plans multiple candidate regions of interest, verifies them through branch-wise local inspection, and maintains an explicit evidence state for inspected locations, verified findings, and unresolved sub-goals. Third, the model needs optimization signals that reward not only final answers but also valid and state-consistent exploration. We develop a GRPO-based reinforcement learning strategy with step-wise rewards for format validity, answer correctness, localization quality, and plan execution consistency.

Overall, GeoVista shifts UHR remote sensing interpretation from passive visual intake and single-path zooming toward planning-driven, evidence-grounded active perception. By combining global scene awareness, branch-wise local verification, and explicit evidence tracking, GeoVista can acquire and aggregate visual evidence across spatially distributed regions more effectively. The main contributions of this paper are summarized as follows:

\begin{itemize}
    \item We construct \textbf{APE-GRO}, a supervised trajectory dataset with 33,445 validated tool-interleaved trajectories for Global-Region-Object active perception in UHR remote sensing.

    \item We propose \textbf{GeoVista}, a planning-driven active perception framework that replaces single-path zooming with multi-branch ROI planning, local verification, and evidence-state tracking.

    \item We develop a two-stage training strategy that combines supervised cold-start fine-tuning with GRPO-based agentic alignment for state-aware evidence acquisition.

    \item Experiments on RSHR-Bench, XLRS-Bench, and LRS-VQA show that GeoVista improves UHR remote sensing understanding through adaptive multi-region evidence aggregation.
\end{itemize}

\section{Related Work}
In this part, we briefly overview the development process of remote sensing vision-language models, tool-interleaved visual reasoning, and active perception methods for UHR remote sensing.

\textbf{Remote Sensing Vision-Language Models.}
Recent multimodal large language models (MLLMs), including GPT-4~\cite{achiam2023gpt}, Gemini~\cite{team2023gemini}, LLaVA~\cite{Liu2023VisualIT,Liu2023ImprovedBW,Li2024LLaVAOneVisionEV}, and Qwen-VL~\cite{bai2023qwenvlversatilevisionlanguagemodel,Wang2024Qwen2VLEV,Bai2025Qwen25VLTR}, have demonstrated strong vision-language understanding capabilities. To adapt these models to aerial and satellite imagery, remote sensing VLMs align RS-specific visual representations with large language models, leading to systems such as RSGPT~\cite{Hu2023RSGPTAR}, SkyEyeGPT~\cite{Zhan2024SkyEyeGPTUR}, GeoChat~\cite{Kuckreja2023GeoChatGroundedLV}, EarthMind~\cite{shu2025earthmindleveragingcrosssensordata}, and EarthVL~\cite{Wang2026EarthVLAP}. Nevertheless, most RS-VLMs still rely on one-shot visual encoding, where a fixed token budget must trade off global coverage and local detail. As a result, a single compact observation is often insufficient for sparse and spatially distributed evidence in UHR scenes. To improve interpretability and performance on complex tasks, recent studies have introduced reasoning paradigms such as chain-of-thought into remote sensing. RS-Thinker~\cite{liu2025towards}, GeoVLM-R1~\cite{fiaz2025geovlm}, GeoZero~\cite{wang2025geozero}, and Geo-R1~\cite{xu2025geo} further emphasize process-level modeling and expose intermediate reasoning steps rather than answering directly. These methods improve complex-task performance and output readability to some extent, but a single round of elaborate reasoning over a fixed visual input is still inadequate when evidence must be acquired across multiple regions and scales. UHR interpretation therefore requires repeated observation and reasoning rather than either one-shot encoding or one-pass textual deliberation alone.

\textbf{Tool-Interleaved Visual Reasoning.}
Recent studies have explored tool-interleaved visual reasoning, where models acquire additional visual evidence during inference rather than relying only on the initial image representation. The ``Thinking with Images'' paradigm~\cite{Su2025ThinkingWI} extends textual Chain-of-Thought reasoning with intermediate visual perception steps. General-domain systems such as DeepEyes~\cite{zheng2025deepeyes}, SenseNova-MARS~\cite{chng2025sensenova}, Vision-R1~\cite{huang2025vision}, and MedVR~\cite{jiang2026medvr} further show that visual tool use and reinforcement learning can improve multi-step perception and evidence grounding. Despite these advancements, general-domain benchmarks such as VisReason~\cite{li2025visreason} and synthetic datasets such as MM-Adaptive-CoF~\cite{zhang2025adaptive} are inherently misaligned with the spatial complexities and scale imbalance of RS imagery. In the RS domain, a significant bottleneck remains the scarcity of specialized interleaved reasoning datasets. Existing RS VQA datasets such as RSVQA~\cite{lobry2020rsvqa} and EarthVQA~\cite{wang2024earthvqa} mainly provide single-turn answer supervision and offer limited guidance on how to search, zoom, and verify evidence across scales in UHR scenes, where relevant evidence may be small, widely separated, and scale-dependent.

\textbf{Active Perception for UHR Remote Sensing.}
Understanding ultra-high-resolution remote sensing imagery remains a bottleneck for vision-language models. GeoLLaVA-8K~\cite{Wang2025GeoLLaVA8KSR} recently extends static encoding to roughly 8K-resolution inputs, but a fixed visual context still faces inherent limits in balancing global coverage and fine-grained local evidence. Active perception offers a complementary direction by progressively acquiring task-relevant observations during inference, making it possible to extend effective resolution beyond a single encoding pass. In remote sensing, ZoomEarth~\cite{liu2025zoomearth} and GeoEyes~\cite{wang2026geoeyes} demonstrate the effectiveness of interactive observation for UHR image interpretation. Related trajectory datasets, including LRS-GRO~\cite{liu2025zoomearth} and UHR-CoZ~\cite{wang2026geoeyes}, introduce multi-level annotations and interleaved reasoning trajectories for active perception. However, existing approaches usually organize exploration as a single sequential path. This makes it difficult to compare multiple candidate ROIs, maintain global context, track verified evidence, and avoid missed or duplicated detections. In contrast, GeoVista introduces planning-driven multi-branch exploration through an Observe-Plan-Track mechanism, enabling global planning, branch-wise local verification, and evidence-state aggregation for UHR remote sensing understanding.

\section{Methodology}
\label{sec:methodology}

Given an UHR remote sensing image $I$ and a user query $Q$, our goal is to generate an answer $A$ by actively acquiring and aggregating visual evidence across multiple spatial scales. Instead of relying on a single global encoding or following a single zooming trajectory, GeoVista formulates UHR interpretation as a planning-driven active perception process. The model first observes the global scene, then plans multiple candidate ROIs, verifies them through branch-wise local inspection, and finally aggregates the verified evidence for prediction. This section introduces the proposed method in three parts. First, we describe \textbf{APE-GRO}, a planning-and-execution trajectory dataset for active perception, in Section~\ref{sec:ape_gro}. Next, we present the \textbf{GeoVista} framework, including the Observe-Plan-Track mechanism for global observation, multi-ROI planning, branch-wise verification, and evidence-state tracking in Section~\ref{sec:geovista}. Finally, we introduce the two-stage training strategy that combines supervised cold-start fine-tuning with GRPO-based alignment in section~\ref{sec:training_pipeline_grpo}. 

\subsection{Cross-Scale Interleaved Trajectory Dataset}
\label{sec:ape_gro}

\begin{figure*}[!t]
    \centering
    \includegraphics[width=\textwidth]{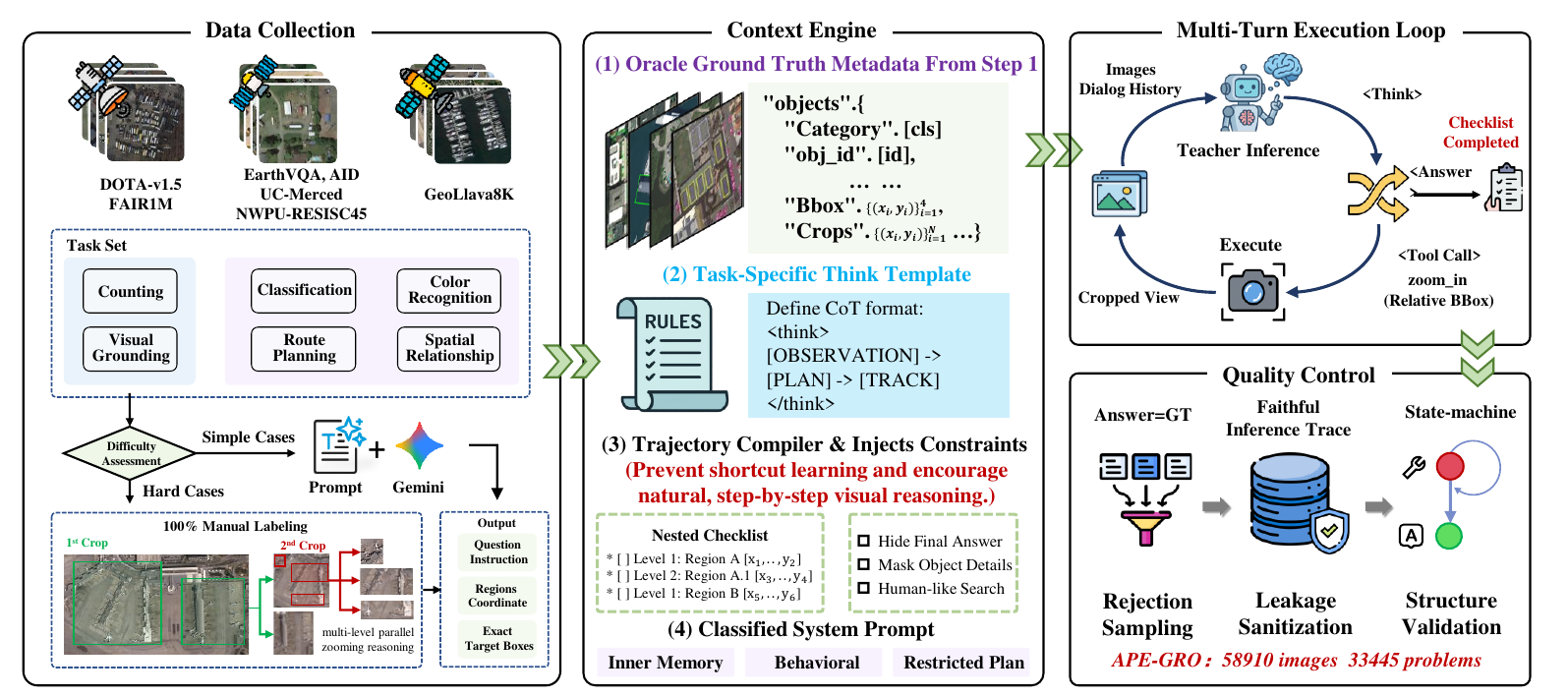}
    \caption{Data construction pipeline for APE-GRO. The process consists of four stages: data collection, context construction, multi-turn execution, and quality control.}
    \label{fig:data_construction}
\end{figure*}

UHR remote sensing interpretation requires models to acquire visual evidence through an explicit search process rather than relying only on a single image-question-answer pair. For tasks involving sparse targets, distributed regions, or fine-grained local details, the model needs to plan where to inspect, execute visual tools, update intermediate evidence, and decide whether further exploration is required. However, most existing RS VQA datasets provide only final-answer supervision and offer limited guidance on how cross-scale evidence should be collected. To provide cold-start supervision for this behavior, we construct \textbf{APE-GRO}, a supervised trajectory dataset for tool-interleaved UHR reasoning that captures \textbf{A}ctive \textbf{P}lanning and \textbf{E}xcution across \textbf{G}lobal, \textbf{R}egion, and \textbf{O}bject levels.

\subsubsection{Trajectory Format}
Each trajectory in APE-GRO is represented as a sequence of reasoning states, tool actions, visual observations, evidence updates, and the final answer:
\begin{equation}
\tau = (z_0, a_0, o_1, z_1, a_1, o_2, \dots, z_T, A),
\end{equation}
where $z_t$ denotes the textual reasoning state, $a_t$ denotes the visual action, $o_t$ is the returned visual observation, and $A$ is the final answer. Each visual action is defined as:
\begin{equation}
a_t = \text{crop}(b_t,\ell_t),
\end{equation}
where $b_t$ is the selected ROI and $\ell_t \in \{L_0,L_1,L_2\}$ indicates the global, region, or object level. In practice, trajectories are serialized with functional fields such as \texttt{<think>}, \texttt{<tool\_call>}, and \texttt{<answer>}. Within \texttt{<think>}, structured sub-steps including \texttt{[Observation]}, \texttt{[Plan]}, and \texttt{[Track]} describe the current visual evidence, planned inspections, and evidence-state updates. This trajectory format provides supervision not only for tool invocation, but also for planning, local verification, and evidence aggregation.

\subsubsection{Data Construction Pipeline}
As shown in Fig.~\ref{fig:data_construction}, APE-GRO is constructed through a four-stage pipeline.

\paragraph{Data collection.}
We collect diverse RS data sources spanning both high- and low-resolution imagery. High-resolution sources include GeoLLaVA-8K~\cite{Wang2025GeoLLaVA8KSR}, FAIR1M~\cite{sun2022fair1m}, and DOTA-v1.5~\cite{xia2018dota}; low-resolution sources include AID~\cite{xia2017aid}, UC-Merced~\cite{yang2010bag}, NWPU-RESISC45~\cite{cheng2017remote}, WHU-RS19~\cite{dai2010satellite}, and EarthVQA~\cite{wang2024earthvqa}. The resulting pool covers six task categories, from compact scene classification to large-area UHR interpretation. For hard cases that require explicit search supervision, we apply task-specific difficulty screening and manual region annotation. In counting tasks, any sample with more than 10 targets is treated as hard: sparse object boxes alone cannot guide a teacher model to produce natural search trajectories, so we manually annotate coarse-to-fine, multi-level focus regions and record the target count within each box to support multi-round zoom trajectory construction. For visual grounding tasks, we additionally construct hard instances with extremely small target boxes and manually annotate progressive focus regions around them, enabling the teacher to learn step-wise localization rather than one-shot retrieval.

\paragraph{Context engine.}
The context engine converts Step-1 annotations into structured teacher inputs in four steps: compiling oracle metadata with category, \texttt{obj\_id}, bounding boxes, and crop coordinates for downstream validation; defining a unified CoT template, \texttt{[OBSERVATION]}$\rightarrow$\texttt{[PLAN]}$\rightarrow$\texttt{[TRACK]}, within \texttt{<think>}; compiling a nested region checklist and injecting constraints to hide the final answer, mask object details, and enforce progressive human-like search; and assembling a classified system prompt with Inner Memory, Behavioral, and Restricted Plan modules.

\paragraph{Multi-turn execution loop.}
Given the compiled context and source image, a teacher VLM iteratively generates structured reasoning, issues \texttt{zoom\_in} calls with relative bounding boxes, receives cropped visual feedback, and updates the nested checklist until exploration is complete. The resulting dialogues are serialized into tool-interleaved trajectories with \texttt{<think>}, \texttt{<tool\_call>}, and \texttt{<answer>} fields.

\paragraph{Quality control.}
We filter generated trajectories through three checks: rejection sampling retains only answer-consistent samples; leakage sanitization removes trajectories with pre-action coordinate injection or premature result disclosure, preserving faithful inference traces; and structure validation applies a state machine to verify multi-turn dialogue format legality.

\begin{figure*}[!t]
    \centering
    \includegraphics[width=\textwidth]{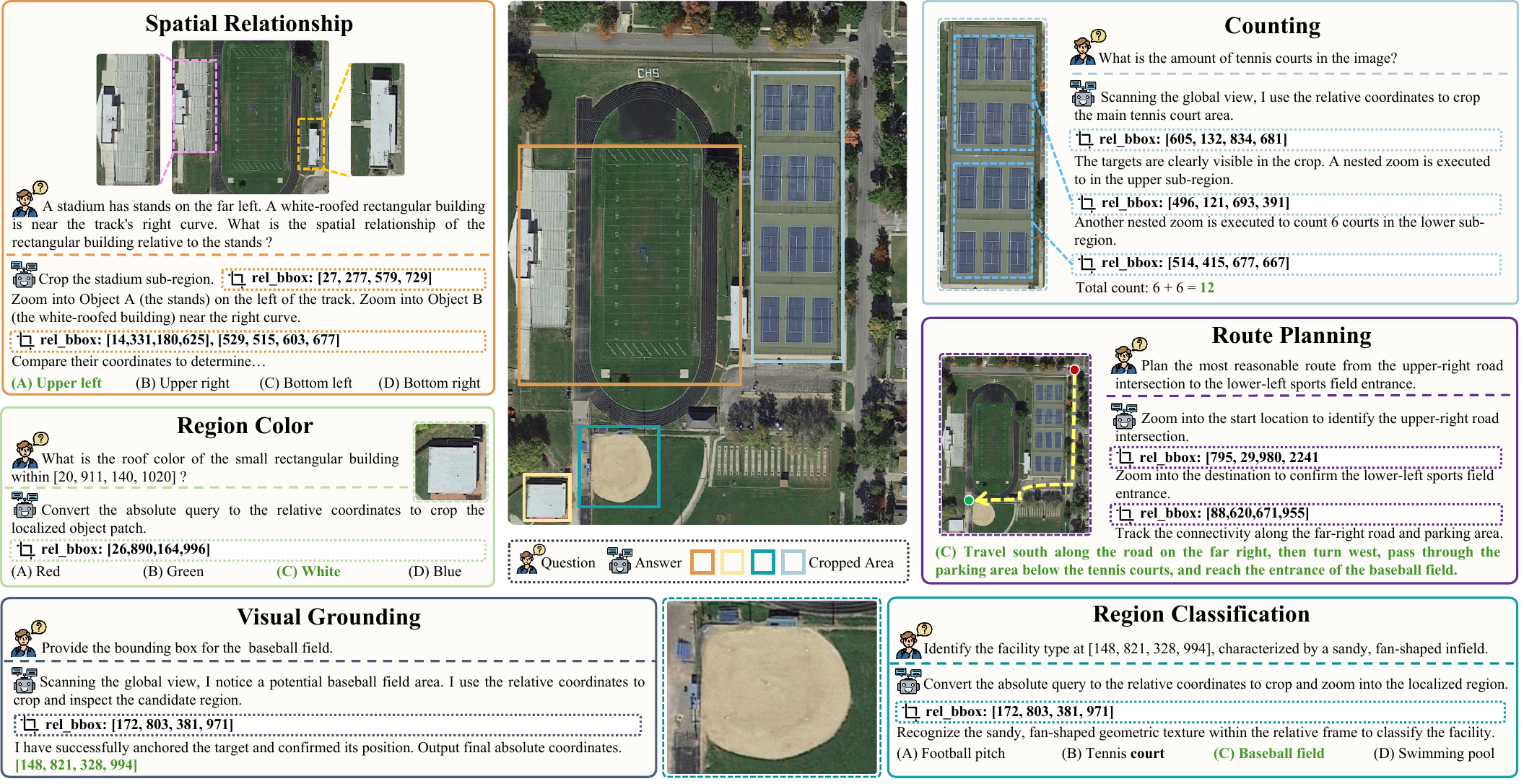}
    \caption{Representative examples from APE-GRO across six UHR task categories: spatial relationship, region color, visual grounding, region classification, route planning, and counting. }
    \label{fig:examples}
\end{figure*}

Starting from 142,125 raw samples, the pipeline produces 33,445 validated trajectories after difficulty-aware labeling, leakage sanitization, structural validation, and answer-consistency checking. Table~\ref{tab:ape_task_dist} summarizes the final task distribution. Representative examples from six UHR task categories are shown in Fig.~\ref{fig:examples}, including spatial relationship, region color, visual grounding, region classification, route planning, and counting. Table~\ref{tab:dataset_comparison} compares our dataset with existing datasets and benchmarks, underscoring its competitiveness in terms of interaction turns, data volume, annotation methods, and its distinctive emphasis on active perception-oriented VQA.

\begin{table}[!t]
    \centering
    \caption{Training sample distribution of APE-GRO across task types and Global, Region, and Object levels.}
    \label{tab:ape_task_dist}
    \resizebox{\linewidth}{!}{
    \begin{tabular}{l|rrrr}
        \toprule
        \textbf{Task Type} & \textbf{Global} & \textbf{Region} & \textbf{Object} & \textbf{Total} \\
        \midrule
        Counting             & 3,615 & 2,059 & 986 & 6,660 \\
        Visual Grounding     & 69    & 1,578 & 20  & 1,667 \\
        Spatial Relationship & 20    & 4,972 & 25  & 5,017 \\
        Route Planning       & --    & 4,875 & 101 & 4,976 \\
        Classification       & 6,497 & 3,703 & --  & 10,200 \\
        Color                & --    & 4,925 & --  & 4,925 \\
        \midrule
        \textbf{Total}       & \textbf{10,201} & \textbf{22,112} & \textbf{1,132} & \textbf{33,445} \\
        \bottomrule
    \end{tabular}}
\end{table}

\begin{table}[!t]
  \centering
  \caption{Comparison with representative remote sensing vision-language datasets and benchmarks. 
  AP denotes whether the dataset involves active perceptual operations, such as cropping, zooming, or region inspection.}
  \label{tab:dataset_comparison}
  \resizebox{\linewidth}{!}{
  \begin{tabular}{lccccc}
    \toprule
    \textbf{Dataset} & \textbf{Avg. Turns} & \textbf{AP} & \textbf{Reasoning Paradigm} & \textbf{\#VQA} & \textbf{Annotation} \\
    \midrule
    \multicolumn{6}{l}{\textit{Passive single-turn benchmarks}} \\
    XLRS-Bench~\cite{wang2025xlrs}      & 1.00 & \xmark & Single-step  & 32,389 & Semi-auto. \\
    RSHR-Bench~\cite{dang2025benchmark} & 1.00 & \xmark & Single-step  & 8,277  & Semi-auto. \\
    LRS-VQA~\cite{luo2025large}         & 1.00 & \xmark & Single-step  & 7,333  & Fully-auto. \\
    \midrule
    \multicolumn{6}{l}{\textit{Active single-path datasets}} \\
    LRS-GRO~\cite{liu2025zoomearth}      & 1.06 & \cmark & Single-path  & 13,245 & Semi-auto. \\
    UHR-CoZ~\cite{wang2026geoeyes}       & 2.15 & \cmark & Single-path  & 25,467 & Fully-auto. \\
    \midrule
    \multicolumn{6}{l}{\textit{Our active multi-path dataset}} \\
    \textbf{APE-GRO}                    & \textbf{2.06} & \cmark & \textbf{Multi-path} & \textbf{33,445} & Semi-auto. \\
    \bottomrule
  \end{tabular}}
\end{table}

\subsubsection{Cross-Scale Organization}
APE-GRO organizes UHR active perception into a three-level Global-Region-Object hierarchy. At the \textbf{Global} level ($L_0$), the model observes a downsampled global view $V^0_{\text{global}}$ and produces a coarse exploration plan. At the \textbf{Region} level ($L_1$), it inspects candidate ROIs $V^1_{\text{crop}}$ to verify structural patterns and narrow down task-relevant areas. At the \textbf{Object} level ($L_2$), it performs fine-grained inspection $V^2_{\text{crop}}$ for small targets, object attributes, and precise localization. This hierarchy encourages global coverage before local evidence acquisition and helps reduce unnecessary inspection of irrelevant regions.

To support consistent tool invocation across different crop sizes and image resolutions, APE-GRO adopts a scale-invariant spatial representation. Absolute pixel coordinates are sensitive to image size, while unrestricted continuous coordinates are difficult to serialize reliably in language-model outputs. We therefore use a discrete relative coordinate system. For a crop ROI $b_i=(x_{\min},y_{\min},w,h)$, each point $(x,y)$ is mapped into a normalized $0$--$1000$ coordinate space:
\begin{equation}
\begin{aligned}
\mathcal{T}(x,y|b_i)
&=
\left(
\mathrm{round}\!\left(\frac{x-x_{\min}}{w} \times 1000\right),
\right. \\
&\quad \left.
\mathrm{round}\!\left(\frac{y-y_{\min}}{h} \times 1000\right)
\right).
\end{aligned}
\end{equation}
For a bounding box $B=(x_1,y_1,x_2,y_2)$, the transformation is applied to its two corner points, i.e.,
$\mathcal{T}(B|b_i)=(\mathcal{T}(x_1,y_1|b_i),\mathcal{T}(x_2,y_2|b_i))$.
This representation provides a nominal relative resolution of $0.1\%$ within each ROI and decouples spatial outputs from the absolute resolution of the original image. 

% \begin{figure*}[!t]
%     \centering
%     \includegraphics[width=\textwidth]{Figures/counting.pdf}
%     \caption{Execution trajectory for the object counting task in APE-GRO.}
%     \label{fig:counting}
% \end{figure*}

\subsection{GeoVista}
\label{sec:geovista}

GeoVista follows the same action space and spatial syntax as APE-GRO during inference. Given an UHR remote sensing image $I$ and a user query $Q$, GeoVista aims to generate an answer $A$ by actively acquiring visual evidence from multiple candidate regions. Instead of encoding the whole image once or greedily following a single zooming path, GeoVista first builds a global exploration plan, then verifies planned ROIs through local observations, and finally aggregates the verified evidence for prediction.

Although all tool calls are decoded sequentially in one autoregressive trace, they are conditioned on a shared global plan and correspond to different planned ROIs. We therefore refer to this process as \textit{multi-branch exploration}: the model does not simply continue along one depth-first zooming path, but maintains multiple candidate regions and updates their evidence states during inference.

\subsubsection{Observe: Global Scene Observation}
GeoVista first obtains a downsampled global view $V^0_{\text{global}}$ from the UHR image. This global view provides broad spatial coverage and supports reasoning about scene layout, potential target distribution, and task-relevant search regions. It is not expected to resolve all fine-grained details. Instead, it serves as the context for subsequent multi-ROI planning.

\subsubsection{Plan: Multi-ROI Exploration Planning}
Conditioned on the query $Q$ and the global view $V^0_{\text{global}}$, the policy $\pi_\theta$ generates an initial structured plan:
\begin{equation}
P^0 = (p_1^0, p_2^0, \dots, p_K^0),
\end{equation}
where each plan item is defined as:
\begin{equation}
p_i^0 = (r_i^0, g_i^0, s_i^0).
\end{equation}
Here, $r_i^0$ denotes the candidate ROI, $g_i^0$ denotes the verification goal for this ROI, and $s_i^0$ denotes its current status. The number of planned ROIs $K$ is constrained by the available tool-call budget. This plan specifies which regions should be inspected and what evidence should be verified in each region, enabling targeted exploration instead of unstructured zooming.

\subsubsection{Track: Branch-wise Verification and Evidence Update}
For each pending ROI $r_i^d$ at depth $d$, GeoVista extracts a high-resolution crop from the original image:
\begin{equation}
v_i^d = \operatorname{Crop}(I, r_i^d).
\end{equation}
The model then decodes a branch trajectory conditioned on the query, global context, current plan, evidence state, selected ROI, and local crop:
\begin{equation}
\tau_i^d \sim \pi_\theta
\left(
\cdot \mid Q, V^0_{\text{global}}, P^d, E, r_i^d, v_i^d
\right).
\end{equation}
After each branch is executed, the evidence state is updated as:
\begin{equation}
E \leftarrow \operatorname{Update}(E, \tau_i^d).
\end{equation}

Each evidence entry records the inspected ROI, its global coordinates, verification status, local observations, detected objects or counts, and unresolved sub-goals. Local crop coordinates are mapped back to the global image frame before being written into $E$, which helps reduce redundant exploration and duplicated counting. If a crop still contains dense or ambiguous content, GeoVista can append a sub-plan $P_i^{d+1}$ and continue inspection at a finer scale. The process terminates when all planned ROIs are verified, the maximum depth is reached, or the tool-call budget is exhausted.

\subsubsection{Evidence Aggregation}
After branch-wise verification, GeoVista aggregates the evidence state to generate the final prediction:
\begin{equation}
A \sim \pi_\theta
\left(
\cdot \mid Q, V^0_{\text{global}}, P, E
\right).
\end{equation}
The aggregation step maps local findings back to their corresponding nodes in the global plan, forming a hierarchical evidence tree that contains verified observations, rejected hypotheses, and unresolved cases. Therefore, the final answer is produced from organized spatial evidence rather than from an unstructured long reasoning history.

\subsection{Training via Supervised Cold-Start and GRPO Alignment}
\label{sec:training_pipeline_grpo}

\begin{figure*}[!t]
    \centering
    \includegraphics[width=\textwidth]{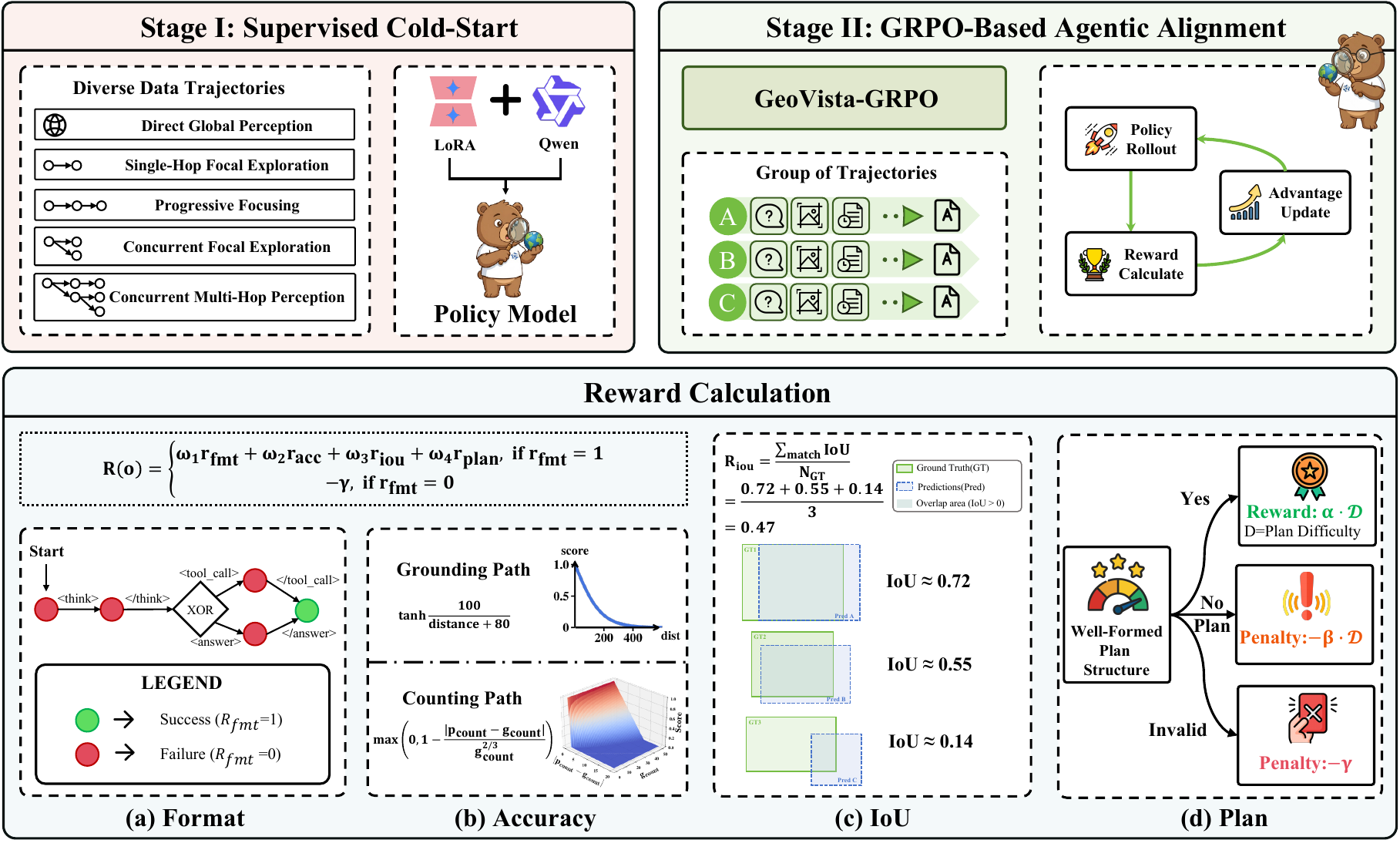}
    \caption{Training pipeline of GeoVista. Stage I performs supervised fine-tuning on APE-GRO trajectories. Stage II applies GRPO-based alignment on verifiable grounding and counting tasks. The plan reward in the figure is a simplified illustration; the exact state-machine formulation is described in the text.}
    \label{fig:training_pipeline}
\end{figure*}

As illustrated in Figure~\ref{fig:training_pipeline}, GeoVista is trained through a two-stage procedure. The first stage provides a supervised cold start for executable active-perception trajectories, while the second stage further aligns the model with task success through verifiable reinforcement learning. This design separates trajectory acquisition from policy optimization: supervised fine-tuning teaches the model to follow the required interaction protocol, whereas GRPO encourages the model to discover more effective exploration and verification strategies in unseen ultra-high-resolution scenes.

\subsubsection{Stage I: Supervised Cold-Start}
Given the trajectory corpus $\mathcal{D}_{\mathrm{SFT}}=\{(I,Q,\tau)\}$, where $I$ denotes the input image, $Q$ denotes the user query, and $\tau$ denotes the reference active-perception trajectory, each trajectory is serialized as an ordered token sequence:
\begin{equation}
    \tau=\{z_t\}_{t=1}^{T},
\end{equation}
where $z_t$ may correspond to reasoning text, plan items, tool calls, evidence updates, spatial coordinates, or final answers. The trajectory therefore provides not only the final response but also the intermediate decision process for coarse-to-fine visual exploration. We optimize the model by maximum likelihood:
\begin{equation}
    \mathcal{L}_{\mathrm{SFT}}
    =
    - \sum_{(I,Q,\tau) \in \mathcal{D}_{\mathrm{SFT}}}
    \sum_{t=1}^{|\tau|}
    \log \pi_\theta(z_t \mid I,Q,z_{<t}).
\end{equation}

This stage initializes the model with valid tool syntax, scale-invariant spatial references, and cross-scale reasoning patterns. In particular, the model learns to decompose ultra-high-resolution visual reasoning into global observation, regional inspection, object-level verification, and final answer generation. Such supervised cold-start is essential for stable reinforcement learning, since malformed trajectories cannot be reliably executed or evaluated by the environment.

\subsubsection{Stage II: GRPO-Based Agentic Alignment}
Although SFT enables the model to imitate reference trajectories, it does not directly optimize exploration strategies for unseen scenes. To further align GeoVista with task success, we apply Group Relative Policy Optimization (GRPO)~\cite{shao2024deepseekmath} on verifiable grounding and counting tasks, where the final outputs can be automatically evaluated using bounding boxes or object counts. Although the reinforcement learning stage is conducted on verifiable tasks, the learned active-perception behavior is shared across different tasks through the same observation, planning, tool invocation, evidence update, and answering interface.

For each training instance $x=(I,Q)$, we sample a group of $G$ trajectories $\{o_i\}_{i=1}^{G}$ from the current policy $\pi_\theta$ and compute a scalar reward $R(o_i)$ for each trajectory. Instead of learning a separate value function, GRPO estimates the advantage of each trajectory by normalizing its reward within the sampled group:
\begin{equation}
\small
    \hat{A}_i =
    \frac{
    R(o_i)-\operatorname{mean}(\{R(o_j)\}_{j=1}^{G})
    }{
    \operatorname{std}(\{R(o_j)\}_{j=1}^{G})+\delta
    },
\end{equation}
where $\delta$ is a small constant for numerical stability. The policy is optimized with the clipped GRPO objective:
\begin{equation}
\small
\begin{split}
    \mathcal{J}_{\mathrm{GRPO}}(\theta)
    &=
    \mathbb{E}_{x,\{o_i\}_{i=1}^{G}}
    \Bigg[
    \frac{1}{G}
    \sum_{i=1}^{G}
    \frac{1}{|o_i|}
    \sum_{t=1}^{|o_i|}
    \\
    &\quad
    \min\left(
    r_{i,t}\hat{A}_i,
    \operatorname{clip}(r_{i,t},1-\epsilon,1+\epsilon)\hat{A}_i
    \right)
    \\
    &\quad
    -
    \beta_{\mathrm{KL}}
    D_{\mathrm{KL}}
    \left(
    \pi_\theta(\cdot \mid x,o_{i,<t})
    \|
    \pi_{\mathrm{ref}}(\cdot \mid x,o_{i,<t})
    \right)
    \Bigg],
\end{split}
\end{equation}
where
\begin{equation}
    r_{i,t}
    =
    \frac{
    \pi_\theta(o_{i,t}\mid x,o_{i,<t})
    }{
    \pi_{\theta_{\mathrm{old}}}(o_{i,t}\mid x,o_{i,<t})
    }.
\end{equation}
Here, $\pi_{\theta_{\mathrm{old}}}$ denotes the behavior policy used for trajectory sampling, and $\pi_{\mathrm{ref}}$ is initialized from the SFT checkpoint. The KL penalty is applied at the token level to prevent the aligned policy from drifting excessively from the executable reasoning behavior acquired during supervised cold-start.

\subsubsection{Reward Design}
For a sampled trajectory $o$, the total reward combines format validity, task accuracy, localization quality, and plan-following quality:
\begin{equation}
\small
\label{eq:reward_total}
R(o)=
\begin{cases}
\omega_1 r_{\mathrm{fmt}}
+
\omega_2 r_{\mathrm{acc}}
+
\omega_3 r_{\mathrm{iou}}
+
\omega_4 r_{\mathrm{plan}},
& \text{if } r_{\mathrm{fmt}}=1, \\
-\gamma_{\mathrm{fmt}},
& \text{if } r_{\mathrm{fmt}}=0.
\end{cases}
\end{equation}
We use format validity as a hard gate because downstream rewards are only meaningful when the trajectory can be parsed into valid actions, spatial references, and final predictions. If the generated output violates the required grammar, it cannot be executed by the environment and is therefore assigned a fixed negative reward $-\gamma_{\mathrm{fmt}}$.

\textbf{Format Reward ($r_{\text{fmt}}$).}
The format reward ensures that the generated trajectory can be parsed and executed. We set $r_{\mathrm{fmt}}=1$ only when the output satisfies all required structural constraints, including valid plan items, legal tool calls, normalized coordinates within the $[0,1000]^2$ range, valid evidence-update fields, and a final answer in the specified format. Otherwise, $r_{\mathrm{fmt}}=0$. This reward prevents the model from obtaining high task scores through unparsable or partially malformed outputs.

\textbf{Accuracy Reward ($r_{\text{acc}}$).}
The accuracy reward evaluates the final task output. For counting tasks, we use a smoothed relative-error score:
\begin{equation}
    r_{\mathrm{acc}}^{\mathrm{count}}
    =
    1-
    \tanh
    \left(
    \frac{|\hat{y}-y|}{\max(y,1)}
    \right),
\end{equation}
where $\hat{y}$ and $y$ denote the predicted and ground-truth object counts, respectively. This formulation gives a continuous reward according to counting accuracy and avoids excessive penalty when the ground-truth count is small.

For grounding tasks, we measure the normalized distance between the predicted and ground-truth box centers:
\begin{equation}
    r_{\mathrm{acc}}^{\mathrm{ground}}
    =
    1-
    \tanh
    \left(
    \frac{
    \|c(\hat{B})-c(B)\|_2
    }{1000}
    \right),
\end{equation}
where $\hat{B}$ and $B$ are the predicted and ground-truth boxes in the normalized coordinate space, and $c(\cdot)$ returns the box center. This term provides a smooth correctness signal for final target localization, while the IoU reward below imposes stricter spatial overlap constraints.

\textbf{Localization Reward ($r_{\text{iou}}$).}
For single-target grounding, localization quality is measured by the standard intersection-over-union:
\begin{equation}
    r_{\mathrm{iou}}^{\mathrm{ground}}
    =
    \operatorname{IoU}(\hat{B},B).
\end{equation}
For counting tasks, the model may output a set of predicted candidate boxes $\mathcal{P}=\{\hat{B}_i\}_{i=1}^{|\mathcal{P}|}$ to indicate the localized objects. To encourage the model to count visually grounded objects rather than relying on coarse scene priors, we define a strict inclusion score for each predicted box:
\begin{equation}
\begin{aligned}
    s(\hat{B}_i)
    &=
    \max_{B_j\in\mathcal{G}}
    \operatorname{IoU}(\hat{B}_i,B_j)
    \cdot
    \mathbb{I}\left[c(B_j)\in \hat{B}_i\right] \\
    &\quad \cdot
    \min\left(
    \frac{a(B_j)}{a(\hat{B}_i)+\eta},
    1
    \right).
\end{aligned}
\end{equation}
where $\mathcal{G}$ is the set of ground-truth boxes, $a(\cdot)$ denotes the box area, $\eta$ is a small constant for numerical stability, and $\mathbb{I}[\cdot]$ is the indicator function. The center-inclusion term requires the predicted box to cover a ground-truth target, while the area ratio penalizes artificially enlarged boxes. The counting localization reward is then computed as
\begin{equation}
    r_{\mathrm{iou}}^{\mathrm{count}}
    =
    \frac{1}{\max(|\mathcal{P}|,1)}
    \sum_{\hat{B}_i\in\mathcal{P}}
    s(\hat{B}_i).
\end{equation}
This term encourages the model to produce localized evidence for counted objects and discourages over-expanded regions that trivially cover multiple targets.

\textbf{State-Aware Dynamic Plan Reward ($r_{\mathrm{plan}}$).}
The plan reward encourages the model to plan before acting, follow the plan during exploration, and update the evidence state consistently. Since different tasks require different degrees of active perception, we introduce a task-dependent difficulty coefficient $\mathcal{D}\in[D_{\min},1]$. For counting tasks, the difficulty increases with the number of targets:
\begin{equation}
    \mathcal{D}_{\mathrm{count}}
    =
    \max\left(
    D_{\min},
    1-\exp(-\lambda_c y)
    \right),
\end{equation}
where $y$ is the ground-truth count. For grounding tasks, smaller targets are considered more difficult in the normalized $1000\times1000$ coordinate space:
\begin{equation}
    \mathcal{D}_{\mathrm{ground}}
    =
    \max\left(
    D_{\min},
    \exp(-\lambda_g |B|/10^6)
    \right),
\end{equation}
where $|B|$ denotes the area of the ground-truth box.

We evaluate plan execution using a lightweight state machine. Each plan item corresponds to an intended inspection unit, such as a candidate region, a target category to verify, or a sub-region generated by decomposition. During trajectory execution, each item is assigned a state, either pending or completed. A pending item is considered completed only after the corresponding inspection is performed and the evidence state is updated. It can also be expanded into a sub-plan when the current evidence is ambiguous or incomplete.

Based on the parsed trajectory, we assign a discrete plan-quality score $Q_{\mathrm{plan}}\in\{1,0,-1\}$. Specifically, $Q_{\mathrm{plan}}=1$ indicates that the trajectory follows valid plan execution, where pending items are either completed with evidence or reasonably expanded into sub-plans. $Q_{\mathrm{plan}}=0$ indicates that the trajectory remains executable but bypasses sufficient verification, such as producing an answer with incomplete evidence. $Q_{\mathrm{plan}}=-1$ indicates severe state-machine violations, such as skipping unvisited regions, marking items as completed without inspection, using inconsistent evidence states, or answering before required verification.

The final plan reward is defined as:
\begin{equation}
    r_{\mathrm{plan}}
    =
    \begin{cases}
    \alpha_{\mathrm{plan}} \mathcal{D},
    & \text{if } Q_{\mathrm{plan}} = 1, \\
    -\beta_{\mathrm{plan}} \mathcal{D},
    & \text{if } Q_{\mathrm{plan}} = 0, \\
    -\gamma_{\mathrm{plan}},
    & \text{if } Q_{\mathrm{plan}} = -1.
    \end{cases}
\end{equation}
In this way, structured exploration is rewarded more strongly on difficult examples, while shallow reasoning on simple examples is not over-penalized. Severe state-machine violations receive a fixed negative reward regardless of task difficulty, which preserves the execution fidelity of active-perception trajectories.

\section{Experiments}
\label{sec:experiments}

\subsection{Experimental Setup}
\label{sec:setup}

\subsubsection{Dataset}
GeoVista is trained and evaluated using complementary data sources that support active perception learning and UHR remote sensing understanding.

\textbf{Training data.}
For Stage I supervised fine-tuning,we use our proposed APE-GRO, a tool-interleaved trajectory corpus introduced in Section~\ref{sec:ape_gro}. APE-GRO contains 33,445 validated trajectories across six task categories, including counting, visual grounding, spatial relationship reasoning, route planning, classification, and color recognition. These trajectories are organized into three perception levels: Global, Region, and Object, as summarized in Table~\ref{tab:ape_task_dist}. For Stage II GRPO alignment, we use the verifiable subsets of APE-GRO, namely visual grounding with 1,667 trajectories and object counting with 6,660 trajectories. These subsets are selected because their outputs can be automatically evaluated using bounding boxes or object counts. To prevent evaluation leakage, samples that overlap with the evaluation benchmarks are removed during dataset construction.

\textbf{Evaluation benchmarks.}
We evaluate GeoVista on three public UHR remote sensing benchmarks with complementary task coverage, as shown in Table~\ref{tab:dataset_comparison}. \textbf{XLRS-Bench}~\cite{wang2025xlrs} contains 32,389 samples and focuses on ultra-high-resolution image understanding through eight perception and reasoning sub-tasks. \textbf{RSHR-Bench}~\cite{dang2025benchmark} contains 8,277 samples and provides a broader evaluation protocol covering reasoning, perception, and multi-turn dialogue. \textbf{LRS-VQA}~\cite{luo2025large} contains 7,333 samples and targets open-ended semantic understanding over large-scale remote sensing imagery, with question types such as counting, attribute recognition, and rural or urban classification. Together, these benchmarks assess whether GeoVista can acquire fine-grained visual evidence and generalize beyond the verifiable tasks used during GRPO alignment.

\subsubsection{Baseline Models}
We compare GeoVista against 19 representative baselines summarized in Table~\ref{tab:mllm_comparison}. The compared methods follow the same grouping as the table: (1) \textit{closed-source MLLMs}, including Gemini-3-flash, Claude-Sonnet-4.5, and GPT-4o; (2) \textit{open-source MLLMs}, including LLaVA1.5-7B, LLaVA-UHD-v3, DeepSeek-VL2-small, and GLM-4V-9B; (3) \textit{large-format MLLMs}, including Qwen2.5-VL-7B and InternVL3.5-8B, which serve as strong static-encoding references with enlarged visual context; (4) \textit{remote sensing MLLMs}, including VHM, GeoChat, EarthDial, GeoLLaVA-8K, Coarse-to-fine, and ZoomSearch; (5) \textit{zoom-in remote sensing MLLMs}, including RS-EoT, ZoomEarth, and GeoEyes; and (6) \textit{zoom-in method}, represented by Qwen-Agent. GeoVista is initialized from Qwen2.5-VL-7B and further trained with APE-GRO and GRPO, making Qwen2.5-VL-7B the most direct static baseline. The zoom-in methods are the closest active-perception competitors for UHR evidence acquisition.

For tool-interleaved evaluation, we adopt each zoom-in method's official or default tool-budget setting, while capping the maximum number of interaction turns at 15 for all active-perception methods, including GeoVista. If a sample terminates because the turn limit is reached, we re-run it up to three times; if all three attempts still hit the turn limit, the prediction is counted as incorrect. Static baselines are evaluated with their default single-pass inference settings.

\subsubsection{Evaluation Metrics}
We report benchmark-level accuracy as the primary metric for all three datasets, following the official evaluation protocol of each benchmark unless otherwise specified. For XLRS-Bench and RSHR-Bench, we use the official accuracy scores defined by the corresponding benchmarks and average them over their respective sub-tasks. For the fine-grained analysis on XLRS-Bench in Table~\ref{tab:xlrs_efficiency}, we additionally report Tools/Q and Tok/Turn to characterize the inference behavior of GeoVista. Tools/Q denotes the average number of tool calls per question, while Tok/Turn denotes the average token consumption per reasoning turn. For LRS-VQA, we follow the official evaluation setting~\cite{luo2025large}, where a prediction is considered correct if its WordNet similarity to the ground-truth answer satisfies $\mathrm{sim} > 0.8$~\cite{miller1995wordnet}. We report both the overall average score and the sub-task scores in Table~\ref{tab:lrs_results}. For RSHR-Bench, Table~\ref{tab:xhr_full_final} further provides a detailed breakdown across reasoning, perception, and multi-turn sub-task groups, along with the corresponding group averages.

\subsubsection{Implementation Details}
All models are initialized from Qwen2.5-VL-7B. SFT uses LoRA (rank=16, $\alpha$=32, dropout=0.05) for 3 epochs with a peak learning rate of $5 \times 10^{-5}$, max sequence length 16,384, and visual resolution capped at 1M pixels. GRPO alignment runs for 2 epochs with learning rate $5 \times 10^{-7}$, 8 rollouts per prompt, and up to 17 active perception turns on 4$\times$ NVIDIA H200 GPUs with bfloat16, Flash Attention 2, and DeepSpeed ZeRO-2. Table~\ref{tab:hyperparams} lists the reward weights and plan-reward coefficients.

\begin{table}[!h]
    \centering
    \caption{Training arguments for RL fine-tuning.}
    \label{tab:hyperparams}
    \renewcommand{\arraystretch}{1.1}
    \begin{tabular}{l c l c}
        \toprule
        \textbf{Parameter} & \textbf{Value} & \textbf{Parameter} & \textbf{Value} \\
        \midrule
        $\omega_1$ & 0.1 & $\omega_2$ & 1.0 \\
        $\omega_3$ & 0.8 & $\omega_4$ & 0.9 \\
        $\alpha$ & 1.0 & $\beta$ & 0.2 \\
        $\gamma_{\text{fmt}}$ & 0.8 & $\gamma_{\text{plan}}$ & 0.8 \\
        $\lambda_c$ & 0.15 & $\lambda_g$ & 120.0 \\
        \bottomrule
    \end{tabular}
\end{table}

\subsection{Main Results}
\label{sec:main_results}

\begin{table}[!t]
\centering
\footnotesize
\renewcommand{\arraystretch}{0.85}
\caption{Overall comparison on UHR remote sensing benchmarks. \textbf{RSHR}, \textbf{XLRS}, and \textbf{LRS} denote official accuracy on RSHR-Bench, XLRS-Bench , and LRS-VQA, respectively. The second-best score in each column is \underline{underlined}.}
\label{tab:mllm_comparison}
\resizebox{\columnwidth}{!}{
\begin{tabular}{l c c c}
\toprule
\textbf{VLM} & \textbf{RSHR} & \textbf{XLRS} & \textbf{LRS} \\
\midrule

\multicolumn{4}{l}{\textcolor{gray}{\textit{Closed-source MLLMs}}} \\
Gemini-3-flash\cite{team2023gemini} & 15.29 & 22.11 & 21.08 \\
Claude-Sonnet-4.5\cite{anthropic2025claude4} & 32.77 & 37.63 & 26.31 \\
GPT-4o\cite{achiam2023gpt} & 32.95 & 32.15 & 26.42 \\
\midrule

\multicolumn{4}{l}{\textcolor{gray}{\textit{Open-source MLLMs}}} \\
LLaVA1.5-7B\cite{Liu2023ImprovedBW} & 29.19 & 35.10 & 23.26 \\
LLaVA-UHD-v3\cite{Sun2025LLaVAUHDVP} & 35.85 & 36.53 & 23.58 \\
DeepSeek-VL2-small\cite{wu2024deepseek} & 35.22 & 40.32 & 26.28 \\
GLM-4V-9B\cite{glm2024chatglm} & 32.70 & 39.77 & 26.86 \\
\midrule

\multicolumn{4}{l}{\textcolor{gray}{\textit{Large-Format MLLMs}}} \\
Qwen2.5-VL-7B\cite{Bai2025Qwen25VLTR} & 32.70 & 41.39 & 22.92 \\
InternVL3.5-8B\cite{wang2025internvl3} & 36.85 & 41.72 & 26.77 \\
\midrule

\multicolumn{4}{l}{\textcolor{gray}{\textit{Remote Sensing MLLMs}}} \\
GeoChat\cite{Kuckreja2023GeoChatGroundedLV} & 29.37 & 26.72 & 13.72 \\
EarthDial\cite{soni2025earthdial} & 30.88 & 33.34 & 18.16 \\
GeoLLaVA-8K\cite{Wang2025GeoLLaVA8KSR} & 31.39 & 37.11 & 21.87 \\
Coarse-to-fine\cite{luo2025large} & 26.79 & 35.65 & 25.33 \\
ZoomSearch\cite{Zhou2025LookWI} & 31.51 & 40.68 & 26.29 \\
\midrule

\multicolumn{4}{l}{\textcolor{gray}{\textit{Zoom-In Remote Sensing MLLMs}}} \\
RS-EoT\cite{shao2025asking} & 31.14 & 37.14 & 18.26 \\
ZoomEarth\cite{liu2025zoomearth} & 31.51 & 34.81 & 19.89 \\
GeoEyes\cite{wang2026geoeyes} & \underline{39.75} & \underline{42.34} & \underline{27.53} \\
\midrule

Qwen-Agent\cite{qwen-agent-cookbook} & 31.20 & 38.12 & 21.64 \\
\midrule

 \textbf{GeoVista (Ours)} & \textbf{41.38} & \textbf{50.65} & \textbf{27.68} \\

\bottomrule
\end{tabular}
}
\end{table}

We report the main experimental results on three evaluation benchmarks to assess the overall effectiveness of GeoVista in UHR remote sensing understanding. Table~\ref{tab:mllm_comparison} summarizes the overall comparison on three UHR remote sensing benchmarks. GeoVista achieves the best performance on all three benchmarks, obtaining 41.38 on RSHR-Bench, 50.65 on XLRS-Bench, and 27.68 on LRS-VQA, exceeding the second-best results by 1.63, 8.31, and 0.15 points, respectively. These results demonstrate the overall effectiveness of GeoVista for UHR remote sensing understanding, especially on benchmarks that require fine-grained evidence acquisition.

Compared with general-purpose MLLMs, GeoVista shows consistent advantages across the three benchmarks. For closed-source models, GeoVista outperforms GPT-4o by 8.43 points on RSHR-Bench and 18.50 points on XLRS-Bench, while achieving comparable performance on LRS-VQA. For open-source general models, GeoVista also surpasses DeepSeek-VL2-small by 6.16 points on RSHR-Bench and 10.33 points on XLRS-Bench. These results suggest that general MLLMs, although capable of broad visual-language reasoning, are less effective when UHR scenes require active localization of small, sparse, or spatially dispersed targets. Without an explicit mechanism to select and inspect task-relevant regions, these models must rely on a fixed visual context, which limits their ability to preserve both global scene structure and local visual details.

GeoVista also compares favorably with zoom-in remote sensing models that introduce focused inspection. Among these methods, GeoEyes is the strongest baseline, achieving 39.75 on RSHR-Bench and 42.34 on XLRS-Bench. GeoVista further improves these scores to 41.38 and 50.65, respectively. The larger gain on XLRS-Bench indicates that a single or sequential zoom-in strategy is still insufficient for complex UHR understanding, where evidence may be distributed across multiple distant regions. By maintaining a global exploration state and verifying multiple task-relevant ROIs, GeoVista can better support tasks involving comparison, counting, and spatial reasoning. The relatively small improvement on LRS-VQA suggests that open-ended semantic understanding benefits less from active region verification than tasks requiring explicit visual evidence localization. Section~\ref{sec:fine_grained_analysis} further analyzes these trends at the sub-task level.

\subsection{Fine-Grained Analysis}
\label{sec:fine_grained_analysis}

We further evaluate GeoVista on the sub-tasks of XLRS-Bench, RSHR-Bench, and LRS-VQA to examine where active multi-region exploration yields the largest gains.

\subsubsection{XLRS-Bench}
\label{sec:xlrs_analysis}

\begin{table*}[htbp]
  \centering
  \caption{\textbf{Experimental results on XLRS-Bench.}
  We compare GeoVista with representative static, remote sensing, and zoom-in methods across perception and reasoning categories. The best results are highlighted in \textbf{bold}, and the second-best results are \underline{underlined}. \textbf{Avg.} denotes the official benchmark score.}
  \label{tab:xlrs_results}
  
  \resizebox{\textwidth}{!}{
  \begin{tabular}{l | c c c c | c c c c | c}
    \toprule
    \multirow{2}{*}{\textbf{Method}} & \multicolumn{4}{c|}{\textbf{Perception}} & \multicolumn{4}{c|}{\textbf{Reasoning}} & \multirow{2}{*}{\textbf{Avg.}} \\
    \cmidrule(lr){2-5} \cmidrule(lr){6-9}
    & \textbf{Counting} & \makecell{\textbf{Scene}\\\textbf{Class.}} & \makecell{\textbf{Object Spatial}\\\textbf{Relationship}} & \makecell{\textbf{Object}\\\textbf{Properties}} & \makecell{\textbf{Complex}\\\textbf{Reasoning}} & \textbf{Planning} & \makecell{\textbf{Spatiotemporal}\\\textbf{Reasoning}} & \makecell{\textbf{Anomaly}\\\textbf{Reasoning}} & \\
    \midrule
    \multicolumn{10}{l}{\textcolor{gray}{\textit{Closed-source MLLMs}}} \\
    Gemini-3-flash\cite{team2023gemini}    & 17.50 & 8.33 & 8.40 & 29.52 & 22.50 & 12.00 & 1.67 & 38.00 & 22.11 \\
    Claude-Sonnet-4\cite{anthropic2025claude4}    & 31.25 & 51.00 & 28.60 & 35.30 & 52.00 & 42.00 & 11.67 & 74.00 & 37.63 \\
    GPT-4o\cite{achiam2023gpt}    & 29.51 & 48.55 & 32.35 & 24.78 & 52.60 & 41.24 & 21.85 & 72.06 & 32.15 \\
    \midrule
    \multicolumn{10}{l}{\textcolor{gray}{\textit{Open-source MLLMs}}} \\
    LLaVA1.5-7B\cite{Liu2023ImprovedBW}    & 30.00 & 39.67 & 25.40 & 34.76 & 49.50 & 35.00 & 20.00 & 64.00 & 35.10 \\
    LLaVA-UHD-v3\cite{Sun2025LLaVAUHDVP}    & 27.50 & 49.33 & 20.80 & 36.93 & 48.50 & 29.00 & 26.67 & 74.00 & 36.53 \\
    DeepSeek-VL2-small\cite{wu2024deepseek}    & 41.25 & 51.67 & 31.60 & 36.69 & 60.00 & 37.00 & 46.67 & 69.00 & 40.32 \\
    GLM-4V-9B\cite{glm2024chatglm}    & 34.38 & 53.67 & 31.20 & 36.27 & 66.00 & 25.00 & 36.67 & 72.00 & 39.77 \\
    \midrule
    \multicolumn{10}{l}{\textcolor{gray}{\textit{Large-Format MLLMs}}} \\
    Qwen2.5-VL-7B\cite{Bai2025Qwen25VLTR}    & 36.88 & 45.00 & 34.20 & 39.64 & 60.00 & 46.00 & 20.00 & \underline{74.00} & 41.39 \\
    InternVL3.5-8B\cite{wang2025internvl3}    & 35.63 & 52.67 & \underline{39.00} & 35.66 & 65.50 & \underline{50.00} & 48.33 & 73.00 & 41.72 \\
    \midrule
    \multicolumn{10}{l}{\textcolor{gray}{\textit{Remote Sensing MLLMs}}} \\
    GeoChat\cite{Kuckreja2023GeoChatGroundedLV}    & 23.75 & 31.33 & 23.80 & 22.17 & 39.00 & 49.00 & 35.00 & 56.00 & 26.72 \\
    Earthdial\cite{soni2025earthdial}    & 33.12 & 21.33 & 26.20 & 32.35 & 60.50 & 27.00 & \underline{55.00} & 61.00 & 33.34 \\
    Geollava-8k\cite{Wang2025GeoLLaVA8KSR}    & 30.00 & \underline{54.00} & 24.20 & 32.17 & 62.00 & \textbf{61.00} & 46.67 & 65.00 & 37.11 \\
    Coarse-to-fine\cite{luo2025large}    & 35.00 & 35.33 & 26.80 & 33.37 & \underline{76.00} & 24.00 & \textbf{58.33} & 69.00 & 35.65 \\
    ZoomSearch\cite{Zhou2025LookWI}    & 36.25 & 48.33 & 28.80 & 39.52 & \textbf{81.00} & 33.00 & 45.00 & \textbf{76.00} & 40.68 \\
    \midrule
    \multicolumn{10}{l}{\textcolor{gray}{\textit{Zoom-In Remote Sensing MLLMs}}} \\
    RS-EoT\cite{shao2025asking} & \underline{41.25} & 46.67 & 23.80 & 36.02 & 58.00 & 46.00 & 1.66 & 58.00 & 37.14 \\
    ZoomEarth\cite{liu2025zoomearth}  & 33.12 & 37.67 & 22.60 & 33.73 & 55.50 & 38.00 & 30.00 & 66.00 & 34.81 \\
    GeoEyes\cite{wang2026geoeyes} & 32.50 & 51.67 & 31.80 & \underline{40.72} & 61.50 & 43.00 & 41.67 & 71.00 & \underline{42.34} \\
    \midrule
    \multicolumn{10}{l}{\textcolor{gray}{\textit{Zoom-In Method}}} \\
    Qwen-Agent\cite{qwen-agent-cookbook} & 25.00 & 50.33 & 28.20 & 37.23 & 48.00 & 31.00 & 50.00 & 67.00 & 38.12 \\
    \midrule
     \textbf{GeoVista (Ours)} & \textbf{41.25} & \textbf{54.00} & \textbf{42.40} & \textbf{52.47} & 52.50 & 46.00 & 48.33 & 69.00 & \textbf{50.65} \\
    \bottomrule
  \end{tabular}
  } 
\end{table*}

As shown in Table~\ref{tab:xlrs_results}, GeoVista achieves the highest overall score on XLRS-Bench, with an average accuracy of 50.65. This surpasses the strongest baseline, GeoEyes, by 8.31 absolute points. The improvement is mainly driven by the perception-oriented sub-tasks. GeoVista obtains 41.25 on Counting and 54.00 on Scene Classification, matching the best baseline results in these two categories. It further achieves the best performance on Object Spatial Relationship and Object Properties, with scores of 42.40 and 52.47, respectively. Compared with the strongest baselines in these two categories, GeoVista improves Object Spatial Relationship by 3.40 points and Object Properties by 11.75 points. These results indicate that GeoVista is particularly effective for fine-grained perception in UHR scenes. The large gain on Object Properties suggests that active multi-region exploration helps the model acquire local visual evidence that is easily diluted in a fixed global view. The improvement on Object Spatial Relationship further shows that maintaining a global exploration state is useful for identifying spatially distributed objects and comparing their relative positions.

In contrast, the gains on reasoning-oriented sub-tasks are less consistent. GeoVista achieves competitive results on Planning and Spatiotemporal Reasoning, but it does not outperform the strongest baselines on Complex Reasoning or Anomaly Reasoning. This suggests that the current active perception strategy mainly improves visual evidence acquisition, while high-level reasoning over abstract events or anomalies may still depend on stronger reasoning priors and task-specific knowledge. Overall, the XLRS-Bench results show that GeoVista's advantage is most pronounced when the task requires locating, inspecting, and integrating fine-grained visual evidence across large UHR images.

\subsubsection{RSHR-Bench}
\label{sec:rshr_analysis}

\begin{table*}[!t]
  \centering
  \caption{\textbf{Experimental results on RSHR-Bench.} Tasks are grouped into reasoning, perception, and multi-turn dialogue. Reasoning---AD=Anomaly (single-turn), FP=Future Prediction (multi-image), MRJC=Multi-region Joint Contrast (multi-image), MRJCS=Multi-region Joint Contrast (single-image, multi-box), OSJ=Object State Judgment (single-turn), Avg.=Reasoning average. Perception---COL=Color Detection, DET=Detection, SHP=Shape Recognition, OC=Object Classification, REL=Object Spatial Relationship, OCN=Object Counting, RCN=Regional Counting, OGD=Object Grounding, RG=Regional Grounding, Avg.=Perception average. Multi-turn---MTFP=Future Prediction, MAD=Anomaly, MOSJ=Object State Judgment, MTEM@1=multi-turn dialogue average.}
  \label{tab:xhr_full_final}
  \resizebox{\textwidth}{!}{
  \begin{tabular}{l | ccccc | c | ccccccccc | c | ccc | c}
    \toprule
    \multirow{2}{*}{\textbf{Model}} & \multicolumn{5}{c|}{\textbf{Reasoning}} & \multirow{2}{*}{\makecell{\textbf{Reas.}\\ \textbf{Avg.}}} & \multicolumn{9}{c|}{\textbf{Perception}} & \multirow{2}{*}{\makecell{\textbf{Perc.}\\ \textbf{Avg.}}} & \multicolumn{3}{c|}{\textbf{Multi-turn}} & \multirow{2}{*}{\makecell{\textbf{MTEM}\\ \textbf{@1}}} \\
    \cmidrule(lr){2-6} \cmidrule(lr){8-16} \cmidrule(lr){18-20}
    & \textbf{AD} & \textbf{FP} & \textbf{MRJC} & \textbf{MRJCS} & \textbf{OSJ} & & \textbf{COL} & \textbf{DET} & \textbf{SHP} & \textbf{OC} & \textbf{REL} & \textbf{OCN} & \textbf{RCN} & \textbf{OGD} & \textbf{RG} & & \textbf{MTFP} & \textbf{MAD} & \textbf{MOSJ} & \\
    \midrule
    \multicolumn{21}{l}{\textcolor{gray}{\textit{Closed-source MLLMs}}} \\
   Gemini-3-flash\cite{team2023gemini} & 28.0 & 38.0 & 10.0 & 18.0 & 26.0 & 25.9 & 15.0 & 19.0 & 5.0 & 20.5 & 10.0 & 7.5 & 14.0 & 17.0 & 11.4 & 13.6 & 31.3 & 26.7 & 41.7 & 4.4 \\
    Claude-Sonnet-4\cite{anthropic2025claude4} & 60.0 & 38.0 & 15.0 & \underline{40.0} & 64.0 & 47.3 & 40.5 & 19.0 & 24.0 & 19.0 & 41.5 & 25.5 & 34.0 & 31.0 & \underline{35.7} & 30.4 & \underline{67.3} & \underline{70.0} & 75.0 & 40.4 \\
    GPT-4o\cite{achiam2023gpt} & 68.0 & \textbf{56.0} & 30.5 & 32.0 & 64.0 & \underline{50.1} & 49.5 & 15.0 & 23.0 & 35.5 & 30.5 & 22.5 & \textbf{41.0} & 28.0 & 27.1 & 30.2 & \textbf{72.0} & 70.0 & \textbf{84.1} & \textbf{47.4} \\
    \midrule
    \multicolumn{21}{l}{\textcolor{gray}{\textit{Open-source MLLMs}}} \\
    LLaVA1.5-7B\cite{Liu2023ImprovedBW} & 40.0 & 28.0 & 30.0 & 30.0 & 46.0 & 35.5 & 51.0 & 24.0 & 20.0 & 29.5 & 27.0 & 18.0 & 24.0 & 25.5 & 22.9 & 28.2 & 46.0 & 60.0 & 48.5 & 14.0 \\
    LLaVA-UHD-v3\cite{Sun2025LLaVAUHDVP} & 38.0 & 28.0 & 20.0 & \textbf{44.0} & 44.0 & 36.8 & 45.0 & 31.0 & 27.0 & \underline{51.5} & 46.0 & \textbf{32.0} & 12.0 & 24.5 & 30.0 & 35.7 & 64.7 & 65.0 & 62.1 & 27.2 \\
    DeepSeek-VL2-small\cite{wu2024deepseek} & 36.0 & 28.0 & 25.0 & 28.0 & \textbf{66.0} & 38.2 & 52.0 & 29.0 & 32.0 & 29.0 & \underline{50.5} & 23.0 & 33.0 & 27.0 & 27.1 & 34.7 & 63.3 & 63.3 & 69.7 & 26.3 \\
    GLM-4V-9B\cite{glm2024chatglm} & \textbf{72.0} & 34.0 & 45.0 & 36.0 & 50.0 & 47.7 & 48.5 & 30.0 & 22.0 & 25.5 & 36.5 & 22.5 & 25.0 & 26.0 & 28.6 & 30.3 & 66.7 & 68.3 & 76.5 & 37.7 \\
    \midrule
    \multicolumn{21}{l}{\textcolor{gray}{\textit{Large-Format MLLMs}}} \\
    Qwen2.5-VL-7B\cite{Bai2025Qwen25VLTR} & 44.0 & 34.0 & 45.0 & 38.0 & 62.0 & 44.6 & 39.5 & 33.0 & 23.0 & 35.0 & 40.0 & 19.5 & 33.0 & 23.5 & 25.7 & 30.8 & 60.0 & 61.7 & 56.1 & 24.6 \\
    InternVL3.5-8B\cite{wang2025internvl3} & \underline{70.0} & \underline{50.0} & 35.0 & 40.0 & 60.0 & \textbf{53.2} & 46.5 & \textbf{39.0} & 25.0 & 46.5 & 19.5 & 24.5 & 33.0 & 32.5 & \textbf{47.1} & 34.2 & 66.7 & \textbf{78.3} & \underline{82.6} & \underline{45.6} \\
    \midrule
    \multicolumn{21}{l}{\textcolor{gray}{\textit{Remote Sensing MLLMs}}} \\
    GeoChat\cite{Kuckreja2023GeoChatGroundedLV} & 42.0 & 26.0 & 45.0 & 24.0 & 46.0 & 35.6 & 39.5 & 24.0 & 22.0 & 34.0 & 27.5 & 21.5 & 34.0 & 24.0 & 22.9 & 28.4 & 50.7 & 58.3 & 56.8 & 13.2 \\
    Earthdial\cite{soni2025earthdial} & 56.0 & 38.0 & 40.0 & 38.0 & 56.0 & 46.4 & 37.5 & 22.0 & 28.0 & 28.5 & 27.0 & 17.0 & 34.0 & 32.5 & 28.6 & 28.4 & 60.7 & 55.7 & 68.9 & 29.0 \\
    Geollava-8k\cite{Wang2025GeoLLaVA8KSR} & 46.0 & 50.0 & 40.0 & 40.0 & 54.0 & 46.8 & 45.5 & 28.0 & 27.0 & 22.5 & 10.5 & 38.5 & 34.0 & 27.0 & 27.1 & 28.9 & 60.7 & 50.0 & 74.2 & 26.3 \\
    Coarse-to-fine\cite{luo2025large} & 26.0 & 22.0 & 35.0 & 26.0 & 30.0 & 26.8 & 27.0 & 28.0 & \underline{33.0} & 27.0 & 25.5 & 25.0 & 29.0 & 23.5 & 30.0 & 26.8 & 21.3 & 28.3 & 25.8 & 2.6 \\
    ZoomSearch\cite{Zhou2025LookWI} & 54.0 & 30.0 & \underline{45.0} & 36.0 & \underline{66.0} & 46.4 & 42.0 & 28.0 & 26.0 & 20.5 & 41.5 & 17.5 & \underline{36.0} & 25.5 & 21.4 & 29.1 & 58.7 & 63.3 & 72.0 & 28.1 \\
    \midrule
    \multicolumn{21}{l}{\textcolor{gray}{\textit{Zoom-In Remote Sensing MLLMs}}} \\
    RS-EoT\cite{shao2025asking} & 48.0 & 26.0 & 40.0 & 32.0 & 62.0 & 41.8 & 49.0 & 19.0 & 20.0 & 31.5 & 38.5 & 18.0 & 28.0 & 23.5 & 21.4 & 29.4 & 37.3 & 35.0 & 40.9 & 10.5 \\
    ZoomEarth\cite{liu2025zoomearth} & 54.0 & 30.0 & 45.0 & 36.0 & 66.0 & 46.4 & 42.0 & 28.0 & 26.0 & 20.5 & 41.5 & 17.5 & 36.0 & 25.5 & 21.4 & 29.1 & 58.7 & 63.3 & 72.0 & 28.1 \\
    Geoeyes\cite{wang2026geoeyes} & 64.0 & 42.0 & \textbf{45.0} & 34.0 & 60.0 & 49.6 & \underline{56.5} & 34.0 & 25.0 & 36.0 & \textbf{55.0} & 19.5 & 34.0 & \textbf{36.5} & 32.9 & \underline{38.2} & 64.7 & 60.0 & 81.1 & 38.6 \\
    \midrule
    \multicolumn{21}{l}{\textcolor{gray}{\textit{Zoom-In Method}}} \\
    Qwen-Agent\cite{qwen-agent-cookbook} & 48.0 & 28.0 & 35.0 & 36.0 & 46.0 & 39.1 & 43.0 & 24.0 & 23.0 & 35.0 & 36.0 & 19.5 & 31.0 & 23.5 & 25.7 & 29.9 & 62.7 & 56.7 & 53.0 & 22.8 \\
    \midrule
     \textbf{GeoVista (Ours)} & 54.0 & 18.0 & 40.0 & 32.0 & 52.0 & 39.1 & \textbf{68.0} & \underline{34.0} & \textbf{50.0} & \textbf{60.5} & 28.0 & \underline{26.5} & 31.5 & \underline{34.0} & 32.9 & \textbf{41.8} & 62.7 & 68.3 & 78.8 & 36.8 \\
    \bottomrule
  \end{tabular}
  }
\end{table*}

Table~\ref{tab:xhr_full_final} reports the detailed results on RSHR-Bench. GeoVista achieves the best performance on the perception subset, reaching a perception average of 41.8 and outperforming GeoEyes by 3.6 points. The improvement is particularly pronounced on fine-grained visual understanding tasks. Specifically, GeoVista obtains 68.0 on color detection, 50.0 on shape recognition, and 60.5 on object classification. It also achieves competitive results on detection and grounding-related tasks. These results demonstrate that the proposed active perception process can effectively locate and inspect informative image regions, leading to more reliable visual evidence acquisition in high-resolution remote sensing scenes.

In contrast, the advantage of GeoVista is less significant on reasoning and multi-turn dialogue tasks. Its reasoning average is 39.1, which trails behind several strong general-purpose and remote sensing MLLMs. This is mainly because RSHR-Bench includes tasks such as future prediction, anomaly analysis, and object state judgment, where performance depends not only on localized perception but also on temporal inference, cross-image comparison, and high-level scene reasoning. Nevertheless, GeoVista remains competitive in several multi-turn tasks. For example, it achieves 68.3 on multi-turn anomaly detection and 78.8 on multi-turn object state judgment. These results indicate that GeoVista primarily improves active fine-grained perception, while more sophisticated temporal and long-horizon reasoning mechanisms remain promising directions for future work.

\subsubsection{LRS-VQA}
\label{sec:lrs_analysis}

\begin{table}[!t]
  \centering
  \caption{\textbf{Experimental results on LRS-VQA.} The sub-tasks include Object Count, Object Background (Bg.), Object Category (Cat.), Object Color (Color), Object Shape (Shape), Object Status (Stat.), Reasoning (Reason), and Rural or Urban (Rur./Urb.).}
  \label{tab:lrs_results}
  \renewcommand{\arraystretch}{1.15}
  \setlength{\tabcolsep}{3pt}
  \resizebox{\linewidth}{!}{
  \begin{tabular}{l | cccccccc | c}
    \toprule
    \textbf{Method} 
    & \textbf{Count} 
    & \textbf{Bg.} 
    & \textbf{Cat.} 
    & \textbf{Color} 
    & \textbf{Shape} 
    & \textbf{Stat.} 
    & \textbf{Reason} 
    & \textbf{Rur./Urb.} 
    & \textbf{Avg.} \\
    \midrule
    \multicolumn{10}{l}{\textcolor{gray}{\textit{Closed-source MLLMs}}} \\
    Gemini-3-flash\cite{team2023gemini}   & 3.92  & 10.20 & 16.01 & 36.47 & 24.07 & 14.30 & 18.10 & 39.94 & 21.08 \\
    Claude-Sonnet-4\cite{anthropic2025claude4} & 3.42  & 14.29 & 7.60  & 45.75 & 36.84 & \textbf{20.80} & 21.90 & 53.91 & 26.31 \\
    GPT-4o\cite{achiam2023gpt}  & 14.18 & 18.37 & 18.34 & 47.06 & 20.98 & \underline{14.70} & \underline{21.90} & 50.80 & 26.42 \\
    \midrule
    \multicolumn{10}{l}{\textcolor{gray}{\textit{Open-source MLLMs}}} \\
    LLaVA1.5-7B\cite{Liu2023ImprovedBW} & 9.34  & \underline{19.18} & 16.92 & 42.88 & 9.72  & 10.30 & 20.40 & 52.64 & 23.26 \\
    LLaVA-UHD-v3\cite{Sun2025LLaVAUHDVP} & 18.52 & 10.20 & 13.17 & 45.88 & 9.72  & 6.90  & 17.90 & 53.27 & 23.58 \\
    DeepSeek-VL2-small\cite{wu2024deepseek} & 14.10 & 13.88 & 19.05 & \underline{48.50} & 18.53 & 10.90 & 20.20 & 55.11 & 26.28 \\
    GLM-4V-9B\cite{glm2024chatglm} & 15.76 & 18.78 & \underline{20.47} & 44.18 & 13.11 & 10.70 & 21.10 & \underline{60.78} & 26.86 \\
    \midrule
    \multicolumn{10}{l}{\textcolor{gray}{\textit{Large-Format MLLMs}}} \\
    Qwen2.5-VL-7B\cite{Bai2025Qwen25VLTR} & 9.76  & 13.88 & 15.91 & 44.97 & 10.96 & 12.80 & 15.20 & 52.08 & 22.92 \\
    InternVL3.5-8B\cite{wang2025internvl3}  & \underline{21.93} & 12.24 & 14.59 & 37.25 & 29.04 & 5.90  & 18.30 & 59.27 & 26.77 \\
    \midrule
    \multicolumn{10}{l}{\textcolor{gray}{\textit{Remote Sensing MLLMs}}} \\
    GeoChat\cite{Kuckreja2023GeoChatGroundedLV}  & 0.17  & 16.73 & 14.18 & 14.25 & 8.02  & 8.00  & 18.30 & 30.35 & 13.72 \\
    Earthdial\cite{soni2025earthdial} & 11.34 & 7.35  & 9.93  & 33.20 & 5.42  & 6.40  & 15.10 & 44.97 & 18.16 \\
    Geollava-8k\cite{Wang2025GeoLLaVA8KSR} & 3.75  & 16.73 & 17.43 & 38.30 & 13.56 & 6.90  & 14.90 & 57.11 & 21.87 \\
    Coarse-to-fine\cite{luo2025large} & \textbf{28.86} & 16.73 & 14.59 & 44.84 & 13.33 & 14.00 & 17.40 & 52.88 & 25.33 \\
    ZoomSearch\cite{Zhou2025LookWI} & 14.10 & \textbf{22.45} & \textbf{23.91} & \textbf{48.76} & 7.80  & 11.00 & \textbf{25.40} & 56.95 & 26.29 \\
    \midrule
    \multicolumn{10}{l}{\textcolor{gray}{\textit{Zoom-In Remote Sensing MLLMs}}} \\
    RS-EoT\cite{shao2025asking} & 15.10 & 10.61 & 9.73  & 30.20 & 23.16 & 7.80  & 11.90 & 32.19 & 18.26 \\
    ZoomEarth\cite{liu2025zoomearth} & 9.34  & 6.94  & 11.04 & 29.67 & \underline{37.40} & 9.50  & 15.50 & 39.70 & 19.89 \\
    Geoeyes\cite{wang2026geoeyes} & 11.26 & 9.80  & 13.17 & 47.06 & 36.16 & 14.00 & 17.50 & 58.75 & \underline{27.53} \\
    \midrule
    \multicolumn{10}{l}{\textcolor{gray}{\textit{Zoom-In Method}}} \\
    Qwen-Agent\cite{qwen-agent-cookbook}  & 9.51  & 11.02 & 10.74 & 30.33 & 31.07 & 5.80  & 12.30 & 52.08 & 21.64 \\
    \midrule
     \textbf{GeoVista (Ours)}
                         & 14.51
                         & 9.80  
                         & 12.16  
                         & 39.22 
                         & \textbf{40.34} 
                         & 10.60 
                         & 17.00 
                         & \textbf{62.22} 
                         & \textbf{27.68} \\
    \bottomrule
  \end{tabular}
  }
\end{table}

Table~\ref{tab:lrs_results} presents the detailed comparison on LRS-VQA. GeoVista obtains the highest average score of 27.68, slightly surpassing GeoEyes and outperforming both general-purpose MLLMs and most remote sensing MLLMs. Unlike the results on RSHR-Bench, the gains on LRS-VQA are not concentrated only on localization-related tasks. Instead, GeoVista shows clear advantages in shape recognition and scene-level rural/urban classification, reaching 40.34 and 62.22, respectively. These two categories require the model to capture object morphology and global land-use context, suggesting that the proposed framework can benefit both local attribute recognition and holistic scene understanding.

It is also worth noting that LRS-VQA covers a diverse set of open-ended visual questions, including object counting, background recognition, category identification, status judgment, and reasoning. GeoVista does not achieve the best result on every sub-task, especially on background, category, color, and reasoning questions, where models such as ZoomSearch and GLM-4V-9B remain competitive. Nevertheless, GeoVista achieves the best overall average without being directly optimized on LRS-VQA during reinforcement learning. This indicates that the learned active observation policy is not limited to the training-oriented grounding and counting objectives, but can generalize to broader high-resolution remote sensing VQA scenarios. The results further suggest that improving how a model explores visual evidence can enhance downstream question answering performance, even when the target tasks involve heterogeneous semantic cues.

\begin{figure*}[!t]
\centering
\includegraphics[width=\textwidth]{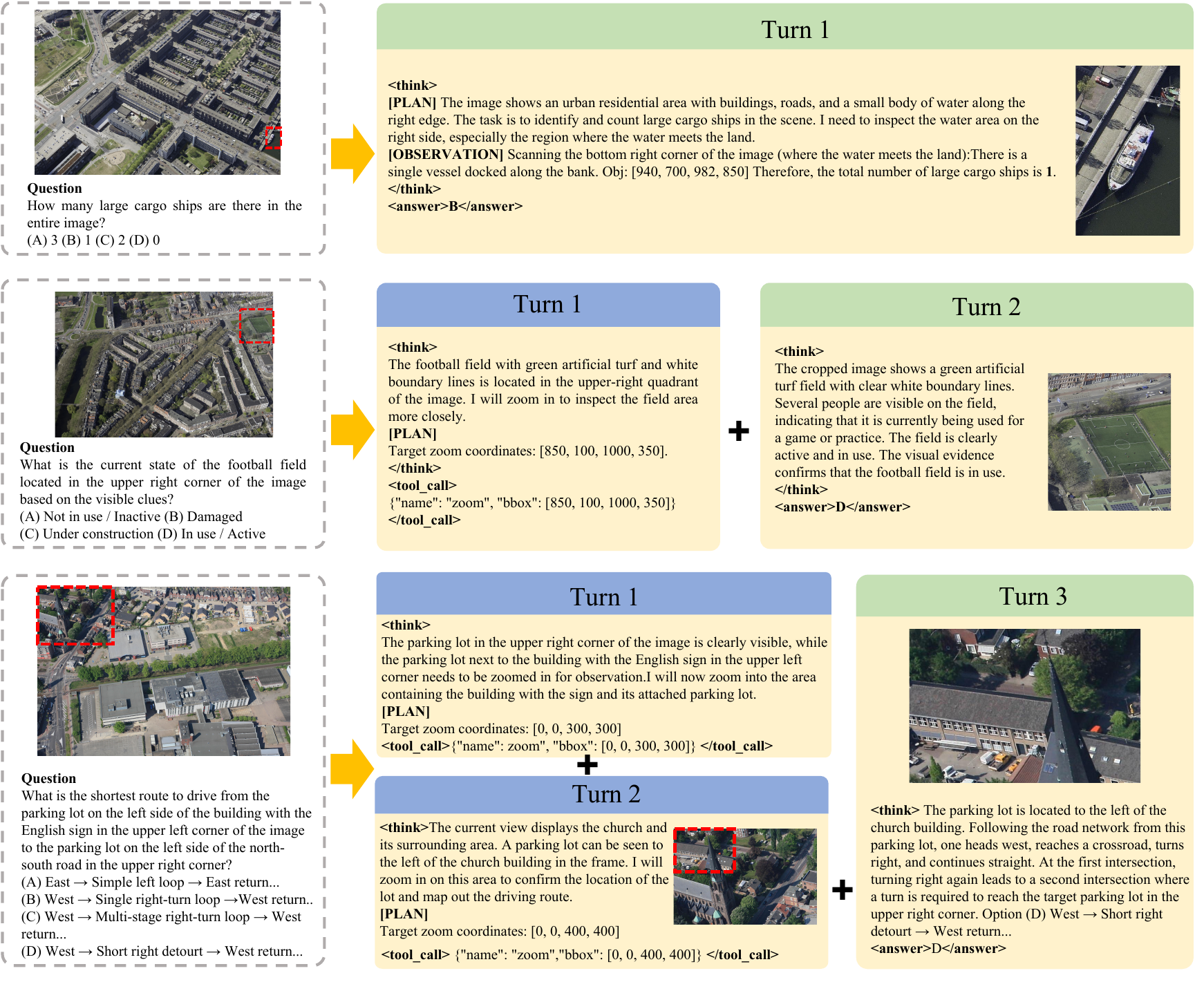}
\caption{Qualitative example of GeoVista for ultra-high-resolution remote sensing image interpretation.}
\label{fig:qualitative_case}
\end{figure*}

\subsubsection{Qualitative Analysis}
As shown in Figure~\ref{fig:qualitative_case}, GeoVista decomposes the interpretation of an ultra-high-resolution remote sensing image into a structured evidence acquisition process. The model first uses the global view to understand the spatial layout of the scene and generate an initial inspection plan. It then zooms into multiple candidate regions to verify local visual details that are difficult to recognize from the global view alone. The inspected regions are recorded in an evidence state and mapped back to the global image coordinates, allowing the model to aggregate observations across spatially separated areas. Compared with single-step visual encoding or single-path zooming, this process provides a clearer connection between the final answer and the visual evidence used for reasoning.

\subsection{Effect of Trajectory Data for SFT}
\label{sec:ablation_sft}

\begin{table}[!t]
    \centering
    \caption{\textbf{Effect of SFT trajectory data.} All variants start from the same Qwen2.5-VL-7B base model and are evaluated after supervised fine-tuning, before RL alignment.}
    \label{tab:sft_ablation}
    \resizebox{\columnwidth}{!}{
    \begin{tabular}{l c c c}
        \toprule
        \textbf{SFT Dataset} & \textbf{RSHR} & \textbf{XLRS} & \textbf{LRS} \\ 
        \midrule
        Base Model\cite{Bai2025Qwen25VLTR}          & 32.70 & 41.82 & 22.92 \\
        + LRS-GRO\cite{liu2025zoomearth}           & 32.83 & 37.42 & 25.08 \\ 
        + UHR-CoZ\cite{wang2026geoeyes}           & 23.02 & 19.44 & 25.72 \\ 
         \textbf{+ APE-GRO} & \textbf{35.15}   {\color{mygreen}\scriptsize$\uparrow$2.45} & \textbf{40.53}  {\color{red}\scriptsize$\downarrow$1.29} & \textbf{27.16}  {\color{mygreen}\scriptsize$\uparrow$4.24} \\
        \bottomrule
    \end{tabular}
    }
\end{table}

Table~\ref{tab:sft_ablation} studies how different trajectory datasets affect the SFT-stage initialization before RL alignment. Compared with the base model, fine-tuning on LRS-GRO slightly improves RSHR-Bench and LRS-VQA, but leads to a clear drop on XLRS-Bench. UHR-CoZ improves LRS-VQA as well, but substantially degrades the performance on both RSHR-Bench and XLRS-Bench. These results indicate that trajectory data with limited interaction diversity or overly sequential zoom-in patterns may improve certain in-domain capabilities, while weakening the model's generalization to broader high-resolution remote sensing scenarios.

In contrast, APE-GRO provides a more balanced and transferable initialization. It improves RSHR-Bench by 2.45 points and LRS-VQA by 4.24 points over the base model, while avoiding the severe performance degradation observed with UHR-CoZ. Although its SFT-only result on XLRS-Bench is slightly lower than that of the base model, the drop is modest. This suggests that APE-GRO does not merely optimize short-term SFT accuracy on a single benchmark. Instead, it mainly equips the model with executable tool-use behaviors, normalized spatial references, and multi-scale Global-Region-Object exploration trajectories, which are important prerequisites for subsequent policy optimization.

This observation is further supported by the RL-stage results. Starting from the APE-GRO SFT checkpoint, GRPO improves the XLRS-Bench score from 40.53 to 50.65. Therefore, APE-GRO should be understood as an RL-ready trajectory initialization rather than a purely performance-oriented SFT dataset. It teaches the model how to conduct structured active perception, while GRPO further optimizes where to inspect, when to continue exploration, and how to aggregate visual evidence for final prediction.

\subsection{Effect of GRPO Alignment}
\label{sec:ablation_rl}

\begin{figure*}[!t]
\centering
\includegraphics[width=\textwidth]{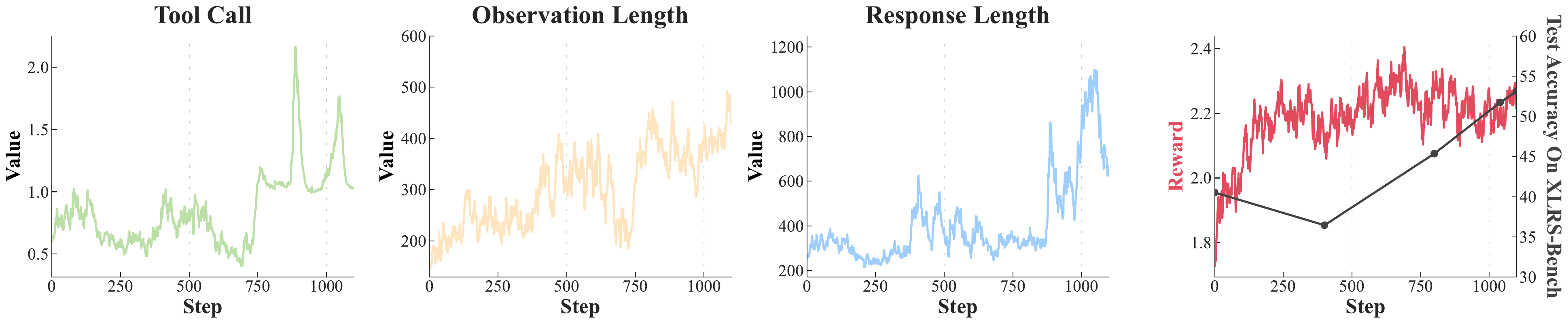}
\caption{Training dynamics during GRPO alignment. We track the aggregated reward, tool usage, observation length, response length, and XLRS-Bench performance across training.}
\label{fig:reward}
\end{figure*}

While SFT teaches the model to imitate structured trajectories, it does not directly optimize exploration decisions in unseen UHR scenes. We therefore analyze how GRPO alignment affects both downstream performance and inference behavior. Starting from the APE-GRO SFT checkpoint, GRPO alignment improves the XLRS-Bench score from 40.53 to 50.65. As shown in Figure~\ref{fig:reward}, the performance does not increase monotonically during training. It first decreases to 36.46 at Step 400, then recovers to 45.36 at Step 800 and reaches 50.65 at Step 1120. This temporary degradation may reflect an early adjustment phase, where the model moves away from supervised trajectory imitation and starts to explore reward-optimized active perception behaviors.

The training curves also reveal a clear change in inference behavior. As GRPO progresses, tool usage and observation length gradually increase, suggesting that the model learns to conduct more active inspections and collect richer visual evidence from UHR images. The later increase in response length further indicates more explicit evidence aggregation after multiple observations. These trends are consistent with the intended Observe-Plan-Track design, where the model is encouraged to inspect multiple informative regions and integrate the collected evidence instead of relying on a single local observation. Notably, although GRPO is trained only with verifiable grounding and counting signals, it improves the overall XLRS-Bench performance, suggesting that reward-aligned active perception behaviors can generalize beyond the directly optimized tasks.

\subsection{Impact of Reward Components}
\label{sec:reward_ablation}

\begin{table}[!t]
\centering
\caption{Impact of different reward components during GRPO alignment. The last row denotes our full method. The last row denotes our full method. $r_{\text{acc}}$, $r_{\text{iou}}$, and $r_{\text{plan}}$ refer to the accuracy, IoU, and state-aware plan rewards, respectively.}
\resizebox{\linewidth}{!}{
\begin{tabular}{cccccc}
\toprule
\multicolumn{3}{c}{\textbf{Reward Components}} & \multicolumn{3}{c}{\textbf{Performance}} \\
\cmidrule(lr){1-3}\cmidrule(lr){4-6}
\textbf{$r_{\text{acc}}$} & \textbf{$r_{\text{iou}}$} & \textbf{$r_{\text{plan}}$} & \textbf{LRS-VQA} & \textbf{RSHR-Bench} & \textbf{XLRS-Bench} \\
\midrule
\xmark & \cmark & \cmark & 25.65 & 38.30 & 42.31 \\
\cmark & \xmark & \cmark & 25.73 & 32.83 & 37.56 \\
\cmark & \cmark & \xmark & 25.71 & 26.67 & 33.47 \\
\midrule
\cmark & \cmark & \cmark & \textbf{27.68} & \textbf{41.38} & \textbf{50.65} \\
\bottomrule
\end{tabular}
}
\label{tab:reward_ablation}
\end{table}

Table~\ref{tab:reward_ablation} presents a leave-one-out ablation study on the reward design used in GRPO. Each variant removes one reward component from the full objective in Eq.~\eqref{eq:reward_total}, while keeping the format reward and the remaining components unchanged. The full reward design achieves the best performance on all three benchmarks, demonstrating that the three reward components are complementary for active perception alignment. Among the three components, the state-aware planning reward plays the most critical role. Removing $r_{\text{plan}}$ causes the largest degradation, reducing the scores to 25.71 on LRS-VQA, 26.67 on RSHR-Bench, and 33.47 on XLRS-Bench. The drop is especially large on XLRS-Bench, where performance decreases by 17.18 points. This indicates that state-aware planning is essential for maintaining structured multi-region exploration and preventing the model from falling into ineffective sequential zooming behaviors. Removing $r_{\text{iou}}$ also leads to clear performance drops, especially on RSHR-Bench and XLRS-Bench, suggesting that explicit spatial overlap supervision helps the model produce more accurate inspected regions and avoid coarse localization. Finally, removing $r_{\text{acc}}$ results in consistent declines across all benchmarks, confirming that final-task correctness remains an important signal for aligning the active perception policy with downstream prediction quality.

\subsection{Efficiency and Cost Analysis}
\label{sec:efficiency_cost}

Table~\ref{tab:xlrs_efficiency} compares zoom-in methods on XLRS-Bench in terms of accuracy, tool usage, and token cost. Qwen-Agent is our Qwen2.5-VL-7B implementation of the Qwen ``Thinking with Images'' paradigm~\cite{Su2025ThinkingWI,qwen-agent-cookbook}, serving as a same-backbone control for GeoVista. 

\begin{table}[!t]
  \centering
  \caption{Accuracy and inference cost of zoom-in methods on XLRS-Bench.}
  \label{tab:xlrs_efficiency}
  \footnotesize
  \renewcommand{\arraystretch}{1.15}
  \setlength{\tabcolsep}{4pt}
  \resizebox{0.8\columnwidth}{!}{
  \begin{tabular}{l | c c c}
    \toprule
    \textbf{Method} & \textbf{Acc.} & \textbf{Tools/Q} & \textbf{Tok/Turn} \\
    \midrule
    Qwen-Agent~\cite{qwen-agent-cookbook} & 38.12 & 0.001 & 54.34 \\
    RS-EoT~\cite{shao2025asking}      & 37.14 & --    & 469.25 \\
    ZoomEarth~\cite{liu2025zoomearth} & 34.81 & 0.82  & 274.30 \\
    GeoEyes~\cite{wang2026geoeyes}    & 42.34 & 0.91  & 253.83 \\
    \midrule
     \textbf{GeoVista (Ours)} & \textbf{50.65} & \textbf{3.82} & \textbf{164.96} \\
    \bottomrule
  \end{tabular}
  }
\end{table}

GeoVista achieves the highest accuracy of 50.65 on XLRS-Bench, outperforming the strongest competing zoom-in method, GeoEyes, by 8.31 points. Compared with prior zoom-in methods, GeoVista invokes tools more frequently, with 3.82 tool calls per question. However, its token cost per turn is substantially lower than RS-EoT, ZoomEarth, and GeoEyes. This indicates that GeoVista does not rely on long single-path reasoning traces. Instead, it performs multiple concise observations over selected regions and aggregates the collected evidence more efficiently.

The comparison with Qwen-Agent further highlights the importance of task-specific active perception training. Although Qwen-Agent uses the same Qwen2.5-VL-7B backbone, it rarely triggers zoom-in operations and achieves only 38.12 accuracy. GeoVista improves over this same-backbone baseline by 12.53 points, demonstrating that the gains do not simply come from the base model capability, but from APE-GRO SFT and GRPO alignment, which teach the model when and where to inspect UHR images.

\section{Conclusion}

In this paper, we introduce GeoVista, an active perception framework for UHR remote sensing imagery. By replacing isolated single-path scanning with Planning-Driven Parallel Multi-Hop Exploration, it achieves seamless global-local integration with high precision. Furthermore, we constructed APE-GRO, a rich interleaved reasoning training set that provides the crucial scale-invariant spatial syntax for initial supervised fine-tuning. We further aligned the agent using a customized GRPO pipeline. By applying multi-dimensional constraints and dynamic rewards, we forced the agent to "think" before acting, successfully internalizing a strict "Observe-Plan-Track" process. Extensive evaluations across RSHR-Bench, XLRS-Bench, and LRS-VQA demonstrate that GeoVista achieves state-of-the-art performance, conclusively proving that structured parallel reasoning and strict step-by-step adherence are the keys to unlocking complex scene understanding in expansive visual environments.

\section*{Acknowledgements}
This work was supported by the National Natural Science Foundation of China under Grant Nos. U22A2098, 62206105, and 62202200; the Scientific and Technological Innovation Project of Changbaishan Laboratory, Jilin Province under Grant No. CBS2026006-04; the China Postdoctoral Science Foundation under Grant No. 2025M780312. This research was also funded by the Major Science and Technology Development Plan of Changchun under Grant No. 2024WX05.

\bibliographystyle{cas-model2-names}

\bibliography{main}

\end{document}